\documentclass[journal,twoside]{IEEEtran}

\usepackage[T1]{fontenc}
\usepackage{amsmath,amsfonts}
\usepackage{amssymb}
\usepackage{algorithmic}
\usepackage{algorithm}
\usepackage{array}
\usepackage[caption=false,font=normalsize,labelfont=sf,textfont=sf]{subfig}
\usepackage{textcomp}
\usepackage{stfloats}
\usepackage{multirow}
\usepackage{cellspace}
\usepackage{float}
\usepackage{url}
\usepackage{verbatim}
\usepackage{tabularx}
\usepackage{tikz}
\usepackage[edges]{forest}
\usepackage{flexisym}
\usepackage{pgfplots}
\usepackage{xcolor}
\usepackage{soul}
\usepackage{filecontents}
\usepackage[normalem]{ulem}
\usepackage{hyperref}
\usepackage{pifont}
\usepackage{caption}
\usepackage{todonotes}
\usepackage{standalone}

\usetikzlibrary{shadows}
\usetikzlibrary{cd}

\graphicspath{ {./figures/} }

\usepackage[backend=biber]{biblatex}
\bibliography{references.bib}

\newcommand*{\etal}{{\textit{et al. }}}

\newcommand{\refinterp}{\hyperref[lab:interpretability]{\textsc{Interpretability}}}
\newcommand{\refguide}{\hyperref[lab:knowledge_guide]{\textsc{Guided Training}}}
\newcommand{\refhetagg}{\hyperref[lab:heterogeneous]{\textsc{Heterogeneous Aggregation}}}
\newcommand{\refunderep}{\hyperref[lab:low_data]{\textsc{Underrepresented Types}}}
\newcommand{\reflongr}{\hyperref[lab:long_range]{\textsc{Long-range Dependencies}}}

\begin{document}

\title{Neurosymbolic AI for Reasoning over Knowledge Graphs: A Survey}

\author{Lauren Nicole DeLong, Ramon Fern\'{a}ndez Mir, Jacques D. Fleuriot
\thanks{This paper was produced by the Artificial Intelligence and its Applications Institute of the School of Informatics at the University of Edinburgh.}%
}

\markboth{}{DeLong, Fern\'{a}ndez Mir \MakeLowercase{\textit{et al.}}: Neurosymbolic AI for Reasoning over KGs}

\maketitle

\begin{abstract}
Neurosymbolic AI is an increasingly active area of research that combines symbolic reasoning methods with deep learning to leverage their complementary benefits. As knowledge graphs are becoming a popular way to represent heterogeneous and multi-relational data, methods for reasoning on graph structures have attempted to follow this neurosymbolic paradigm. Traditionally, such approaches have utilized either rule-based inference or generated representative numerical embeddings from which patterns could be extracted. However, several recent studies have attempted to bridge this dichotomy to generate models that facilitate interpretability, maintain competitive performance, and integrate expert knowledge. Therefore, we survey methods that perform neurosymbolic reasoning tasks on knowledge graphs and propose a novel taxonomy by which we can classify them. Specifically, we propose three major categories: (1) logically-informed embedding approaches, (2) embedding approaches with logical constraints, and (3) rule learning approaches. Alongside the taxonomy, we provide a tabular overview of the approaches and links to their source code, if available, for more direct comparison. Finally,  we discuss the unique characteristics and limitations of these methods, then propose several prospective directions toward which this field of research could evolve. 
\end{abstract}



\section{Introduction}
\IEEEPARstart{K}{nowledge Graphs} (KGs) are becoming an increasingly popular way to represent information because they are multi-relational, easily queried, and can store various types of data in a similar, consistent format \cite{vicknair2010comparison}. Furthermore, numerous methods have been developed to mine information and make novel predictions from KGs \cite{ji2021survey, wu2020comprehensive}, leading to advancements in drug discovery \cite{schultz2021covid, zhang2021drug}, improved user recommendation systems \cite{yang2021consisrec}, and more efficient traffic forecasting \cite{guo2019attention, wu2022traversenet}, amongst other applications. Traditionally, to make novel predictions about a knowledge base, approaches utilized either a set of rules \cite{galarraga2013amie, ott2021safran} or generated numerical representations from which patterns could be extracted \cite{perozzi2014deepwalk, trouillon2016complex}. Recently, however, newer methods for knowledge discovery and reasoning on graphs have attempted to blend these categories into novel neurosymbolic, or hybrid, methods.

Neurosymbolic artificial intelligence (AI), an increasingly  active field of research, describes the combination of symbolic AI, which often includes logic and rule-based approaches, with neural networks and deep learning \cite{garcez2020neurosymbolic, hitzler2022neuro}. Often, neurosymbolic AI's main advantages are described as achieving comparable performance to current deep learning methods while simultaneously fostering inherent interpretability \cite{garcez2020neurosymbolic, tsamoura2021neural, hitzler2022neuro}. However, there are many different ways in which this new field has been approached. \IEEEpubidadjcol With this article, we aim to provide a comprehensive overview and structural classification of neurosymbolic methods designed for reasoning over KGs by describing their:
\begin{itemize}
    \item overarching taxonomy,
    \item benefits and weaknesses,
    \item applications for which they are already used, and
    \item potential research directions.
\end{itemize}

Specifically, we classify approaches into three major categories: (1) logically-informed embedding approaches (\S\ref{subsection:fusion}), which use symbolic inference and deep learning sequentially, (2) approaches which learn embeddings with logical constraints (\S\ref{subsection:logical_constraints}), and (3) those which learn logical rules for reasoning (\S\ref{subsection:learning_rules}). Moreover, by distinguishing five \textit{critical characteristics} of these approaches, namely \mbox{\refinterp}, \mbox{\refguide} via rules and knowledge, the ability to encode \mbox{\refunderep} and \mbox{\reflongr}, as well as the efficient aggregation of heterogeneous information (\mbox{\refhetagg}), we explore how each category of approaches has unique capabilities for various facets of research.

\subsection{Related Work and Novel Contributions}
\label{subsection:related_works}

\noindent This survey is the first to provide a full, comprehensive overview of neurosymbolic methods for reasoning over KGs. Because the topic is still relatively young, earlier articles tend to focus largely on background information. For example, an article by Zhang \etal \cite{zhang2021neural} discusses general neurosymbolic AI, neural-network-based graph reasoning, and symbolic graph reasoning. However, it only covers a subset of neurosymbolic methods for reasoning on KGs. Similarly, a paper by Chen \etal \cite{chen2020review} provides an expansive overview of the information present in our \textit{Background} (\S\ref{section:background}) and \textit{KG Completion} (\S\ref{section:kg_complete}) sections. Previously, a relevant review by Lamb \etal \cite{lamb2020graph} described how several types of graph neural networks already possess tools, such as attention mechanisms, to be considered or used for neurosymbolic purposes. Additionally, work by Boschin \etal \cite{boschin2022combining} overlaps with our paper. However, we take a different approach toward grouping and describing neurosymbolic methods for KGs. In particular, while Boschin \etal focus on grouping methods by the various rules or domain-specific knowledge used, we outline a taxonomy in terms of the structure of the architectures, including the orders in which symbolic and neural modules are placed and integrated.


\section{Background}
\label{section:background}

\subsection{A Brief Introduction to Knowledge Graphs}
\label{subsection:intro_to_kgs}

\noindent A \emph{graph} or \emph{network} structure $G$ is composed of a finite set of vertices or nodes, \emph{V}, and a finite set of edges, \textit{E}, connecting the vertices. A graph $G$ can be represented, then, as a tuple of these two sets, $(V, E)$. Each edge $(u, v) \in E$ connects two \textit{adjacent} or \emph{neighbouring} vertices, $u$ and $v \in V$. The term \textit{triple} denotes a pair of vertices connected by an edge \cite{hamilton2020graph}.

A \textit{knowledge graph} (KG) uses a graph structure to represent a \textit{knowledge base (KB)}, a collection of factual triples denoting relations between real-world entities \cite{ji2021survey}. KGs have been employed in research domains from biochemistry and medical data \cite{domingo2021covid, gaulton2017chembl} to social media platforms \cite{huang2020embedding}. They may be as simple as nodes and edges, or they may additionally contain semantic information, relation and entity types, and node and edge properties \cite{chen2020knowledge, ji2021survey}. KGs have become one of the most popular ways to represent information, as they capture multi-relational data well, are fast and easy to query when stored in a graph database, and allow the representation of many different types of data in a similar format (a ``universal language''). Some of the best known KGs, such as YAGO \cite{suchanek2007yago} and DBpedia \cite{auer2007dbpedia}, contain millions of entities and relations representing general information about people, places, and other general items such as movies and music and are widely used as benchmark datasets. Other smaller KGs exist to store information about niche areas, such as the COVID-19 pandemic \cite{domingo2021covid, schultz2021covid}.

KGs are also becoming an increasingly popular way to represent \textit{growing} or \textit{changing} KBs due to their flexible storage schemas \cite{vicknair2010comparison}. However, because KGs house information that we, as people, know about some given domain, they are inherently incomplete in comparison to the real world. The practice of adding new information to a graph, known as \textit{KG completion}, is, therefore, a common way to make predictions, a topic discussed further within \S\ref{subsection:kg_completion_reasoning}.

\subsection{A Brief Introduction to Logic-Based Reasoning}
\label{subsection:introduction_to_logic}

\noindent While the ``symbolic'' components of neurosymbolic approaches vary, many of them utilize some form of \textit{logic}-based reasoning. We can think of a logic as a formal system comprised of syntax, semantics and inference rules. Different logics may suit different applications either because of their abilities to represent interesting properties or because of their levels of automation. In this section, we will cover logic programming, probabilistic logic programming, fuzzy logic and description logic, highlighting how they are used for reasoning on graph structures.

\emph{Logic programming} \cite{baral1994logic} is one of the most prominent and widely used symbolic reasoning techniques. The fundamental building blocks of this paradigm are \textit{Horn clauses} \cite{horn1951sentences}, which are formulas of the form:
\[
    H \leftarrow B_1 \land \cdots \land B_n.
\]
These are meant to be interpreted as rules where the body $B_1 \land \cdots \land B_n$ implies the (positive) head $H$, or as facts, if the body is empty. In its most basic form, every element is an atom and all the free variables are implicitly universally quantified. For example, consider the following background knowledge and rule:
\begin{align*}
    &\text{likes}(alice, bob), \\
    &\text{likes}(bob, alice), \\
    &\text{friends}(X, Y) \leftarrow \text{likes}(X, Y) \land \text{likes}(Y, X).
\end{align*}
\noindent Intuitively, we can deduce that $\text{friends}(alice, bob)$. 


From this, an interesting question arises: given some background knowledge and a set of positive and negative examples, can we come up with consistent rules? This is the approach in \emph{inductive logic programming} (ILP) \cite{muggleton1994inductive, zhang2023critical}. The First-Order Inductive Learner (FOIL) \cite{quinlan1990learning} is a classical ILP implementation that iteratively creates rules using simple heuristics to pick candidates. The basic approach relies on clean data and a small knowledge base, although there are relevant tools such as Progol \cite{muggleton1995inverse} or Aleph \cite{srinivasan2001aleph} that improve on this classical approach. There are also many variations and extensions of ILP using advanced statistical methods that refine the search procedure, for instance, SHERLOCK \cite{schoenmackers2010learning} and AMIE \cite{galarraga2013amie}, which is mentioned again in \S\ref{subsection:rulebased_kg_completion}. Cropper \etal \cite{cropper2020turning} provide a comprehensive survey of ILP, and Zhang \etal survey its use in explainable AI \cite{zhang2023critical}; we refer the interested reader to those surveys for further information. 


A useful extension to Horn clauses to guide the search is to consider \textit{probabilistic logic} \cite{nilsson1986probabilistic} with \textit{soft rules} \cite{nakashole2012query, guo2018knowledge}, where a probability \(\in [0, 1]\) is attached, \textit{e.g.}
\[
    {0.75\text{ :: car}(X) \leftarrow \text{wheels}(X, 4) \land \text{drivable}(X)}
\]
says that we are 75\% certain that four-wheeled drivable objects are cars. Many tools take this probabilistic approach as it allows noisy data and uncertainty, and takes advantage of statistical methods. A good example is ProbLog \cite{de2007problog} where all the rules are soft and independent, and the resolution algorithm is extended to keep track of probabilities. 

Alternatively, we can attach a weight $w_i$ to every first-order formula $F_i$ in a knowledge base, i.e. $(F_i, w_i)_{i=1}^n$. Assuming that our language has finitely many constants, we can now define a graph called a \textit{Markov Logic Network} (MLN) \cite{richardson2006markov}. First, recall that an \textit{atom} is a formula composed of constants, variables and functions but with no logical connectives, and that a formula is \textit{grounded} if it has no variables. The nodes of a MLN are every grounding of every atom appearing in the $F_i$s, and there is an edge if the ground atoms appear together in at least one of the $F_i$s. We will illustrate the concept by adapting an example from \textit{Domingos et al.} \cite{domingos2008markov}. Suppose we have constants $A$ and $B$ and the following formulas:
\begin{align*}
    &\forall x. \, \text{Vegan}(x) \Rightarrow \text{Healthy}(x) \\ 
    &\forall x. \, \forall y. \, \text{Friends}(x, y) \Rightarrow (\text{Vegan}(x) \Leftrightarrow \text{Vegan}(y)) 
\end{align*}
with weights $1.3$ and $0.7$ respectively.

The ground Markov network looks as follows: 
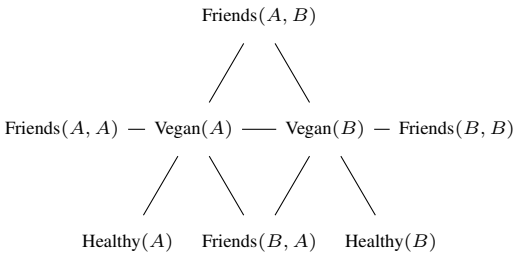
\begin{center}
    \tikzset{main node/.style={minimum height=0.7cm, minimum width=1.25cm, align=center}}
    \begin{tikzpicture}
        \node[main node] (1) at (2.625, 3) {\scriptsize $\text{Friends}(A,B)$};
    
        \node[main node] (2) at (0, 1.5) {\scriptsize $\text{Friends}(A,A)$};
        \node[main node] (3) at (1.75, 1.5)  {\scriptsize $\text{Vegan}(A)$};
        \node[main node] (4) at (3.5, 1.5)  {\scriptsize $\text{Vegan}(B)$};
        \node[main node] (5) at (5.25, 1.5)  {\scriptsize $\text{Friends}(B,B)$};
        
        \node[main node] (6) at (0.875, 0) {\scriptsize $\text{Healthy}(A)$};
        \node[main node] (7) at (2.625, 0) {\scriptsize $\text{Friends}(B, A)$};
        \node[main node] (8) at (4.375, 0) {\scriptsize $\text{Healthy}(B)$};
    
        \draw
        (1) -- (3)
        (1) -- (4)
        (2) -- (3)
        (3) -- (4)
        (4) -- (5)
        (6) -- (3)
        (7) -- (3)
        (7) -- (4)
        (8) -- (4);
    \end{tikzpicture}
\end{center}
A possible \textit{world}, $x$, is a truth value assignment to each node. The probability distribution over the possible worlds, denoted by the random variable $X$, is given by
\[P(X=x) = \frac{\text{exp}\left(\sum_{i=1}^n w_i n_i(x)\right)}{\sum_{y}\text{exp}\left(\sum_{i=1}^n w_i n_i(y)\right)}
\]
where $n$ is the number of formulas and $n_i(x)$ is the number of true groundings of $F_i$ in $x$. In our example, let $x$ be the world where $A$ and $B$ are vegan, unhealthy and have no friends. We can compute that $P(X=x) \approx 0.00055$. With this machinery in place, we can answer queries such as the probability that $A$ is healthy.

Instead of associating a weight (or probability) to each formula, representing the ``degree of belief'', we can also attach a value to the ``degree of truth'' of the statement. This approach is known as \textit{fuzzy logic}, in which the interpretation of the usual boolean logical connectives is adjusted \cite{zadeh1988fuzzy}. For instance, a fuzzy version of $x \land y$ could be $\text{min}(x,y)$ where $x$ and $y$ are in the range $[0,1]$. In general, any  binary operation on $[0,1]$ that is commutative, monotone, associative, and respects the identity is called a \textit{t-norm} and can represent conjunction in fuzzy logic. It is also possible to go back and forth while preserving some of the semantics. Doing so allows us to apply statistical inference to an originally discrete problem and, conversely, logical reasoning to uncertain data. As we will see in \S\ref{rule-based-restrictions-on-possible-predictions}, fuzzy inference systems can be used to increase explainability in deep learning systems \cite{bonanno2017approach}.

Another angle is to consider specific fragments of first-order logic with different levels of expressive power. \emph{Description logics} \cite{krotzsch2012description} are \textit{knowledge representation languages} that have a rich syntax to talk about concepts, hierarchies, and cardinalities. Therefore, they provide a convenient semantic framework to reason over KGs, particularly in an ontological sense. In fact, we can think of ontologies as data schemas for KGs. The W3C Web Ontology Language (OWL) \cite{mcguinness2004owl} is an important implementation of this idea. Originally, OWL was designed to describe the Semantic Web, but now it is integral to several neurosymbolic systems \cite{herron2023benefits}. There are also variations such as OWL 2 EL, designed with biomedical ontologies in mind, which we will see in  \S\ref{subsection:fusion}. Here, EL means ``existential language'', so in OWL 2 EL we can have existential but not universal quantification. These restrictions enable polynomial time reasoning, which is crucial for large knowledge bases. In this context, reasoners are pieces of software that can infer logical consequences from a set of formulas.


\subsection{A Brief Introduction to Neural Networks}
\label{subsection:introduction_to_nns}

\noindent As the name suggests, the ``neural'' components of neurosymbolic approaches comprise neural network-based structures. Recently, much state-of-the-art work in machine learning (ML) and AI involves artificial neural networks (NN) \cite{rosenblatt1958perceptron, tino2015artificial} due to their impressive abilities to learn functions that capture underlying patterns in data \cite{krizhevsky2012imagenet, schlichtkrull2018modeling}. A NN uses a series of nodes, or \textit{neurons}, with activation functions to act as some \textit{firing threshold}. Similarly to biological neurons, if the input into each activation function exceeds that threshold, then there is an output. Specifically, \textit{deep} NNs are those with two or more layers \cite{tino2015artificial, skansi2018introduction}.

In recent years, many deep NN variants have sprouted to handle various data structures and accomplish novel goals. Convolutional NNs (CNNs) \cite{krizhevsky2012imagenet}, for example, are generally used for image data, recurrent NNs (RNNs) with Long Short-Term Memory (LSTM) are ideal for dynamic temporal data \cite{hochreiter1997long, sherstinsky2020fundamentals}, and several forms of graph NNs (GNN) have been designed to handle multi-relational data in graph structures, such as factual triples in KGs, as described previously in \S\ref{subsection:intro_to_kgs} \cite{wu2020comprehensive}. However, many NNs are \textit{black-box} models which lack interpretability \cite{zhang2021survey}. This means that a human cannot inherently understand the internal processes that lead to the model output. With multiple layers and sometimes up to millions or billions of parameters \cite{brown2020language, chowdhery2022palm}, it becomes infeasible for a human to follow the individual actions taken by the model to derive predictions. In some cases, such as biomedical applications in which datasets may contain demographic bias, this may be ethically wrong or even dangerous \cite{olejarczyk2021patient, norori2021addressing}. As a result, many current efforts centre upon adding some aspect of interpretability to such methods \cite{zhang2021survey}.


\subsection{Neurosymbolic AI: A Hybrid Approach}
\label{subsection:neurosymbolic_ai}

\noindent \textit{Neurosymbolic AI} is the field of research that studies the combination of deep learning and symbolic reasoning \cite{tsamoura2021neural, garcez2020neurosymbolic, hitzler2022neuro}. The argument for this hybrid approach is that neural and symbolic systems can complement each other and mitigate their respective weaknesses. For many applications, it is highly desirable to combine accountable and interpretable logic-based modules with effective deep learning ones. While interpretability is defined in various ways across the literature, it is generally viewed as the ability to be understood by a human \cite{molnar2020interpretable, zhang2021survey}. Essentially, 
an end-user might easily comprehend the reasoning processes that led to model predictions.

Based on the classification given by Kautz at AAAI-20\footnote{\url{https://roc-hci.com/announcements/the-third-ai-summer/}} (available as a written summary \cite{kautz2022third}), neurosymbolic integration is often categorized as follows \cite{sarker2021neuro}:

 \subsubsection{\textsc{Symbolic Neuro Symbolic}} a NN takes a symbolic representation as input and reconstructs another symbolic representation as output. 
\subsubsection{\textsc{Symbolic[Neuro]}} systems where a symbolic engine queries a NN during reasoning to, for example, estimate a utility function.
\subsubsection{\textsc{Neuro;Symbolic}} the NN and the symbolic module solve complementary tasks and communicate frequently to guide each other.
\subsubsection{\textsc{Neuro:Symbolic $\rightarrow$ Neuro}} symbolic and neural components are tightly-coupled, but symbolic knowledge is ``compiled'' into the training set.
\subsubsection{\textsc{Neuro}\textsubscript{\textsc{Symbolic}}} symbolic logic rules provide the template for the structure of the NN by representing them as tensor embeddings.
\subsubsection{\textsc{Neuro[Symbolic]}} neural engines capable of logical reasoning at certain points in the execution.

To connect our survey back to this preliminary classification of neurosymbolic systems, we will refer back to which of these categories, if any, each section of approaches is most aligned. Generally, in \S\ref{subsection:fusion} and \S\ref{subsection:learning_rules}, the approaches most resemble the \mbox{\textsc{Neuro:Symbolic $\rightarrow$ Neuro}} and \mbox{\textsc{Neuro;Symbolic}} categories. Systems that train with logical constraints are explained in  \S\ref{subsection:logical_constraints}, and fall mostly into the \mbox{\textsc{Neuro}\textsubscript{\textsc{Symbolic}}} and \mbox{\textsc{Neuro[Symbolic]}} classes.


\section{A History of Knowledge Graph Completion}
\label{section:kg_complete}

\noindent While there are various reasoning tasks which could be executed upon KGs, those which perform KG completion (see below) are especially common in the literature, and, more particularly, in the approaches surveyed in the next sections. Thus, we provide a brief background to understand the main body of this article.

\subsection{KG Completion and Reasoning}
\label{subsection:kg_completion_reasoning}

\noindent Since KGs house human knowledge about some given domain, they are notoriously incomplete. KGs could contain false positives based on misinformation and incorrect assumptions. Furthermore, it is often unclear as to whether the absence of details in the KG should indicate missing information or the negation of such information \cite{chen2020knowledge}. \textit{KG completion} (KGC) denotes a series of methods which can be used to refine the graphs and uncover novel information. For example, such completion can reveal unknown molecular functions in a biological KG \cite{schultz2021covid, domingo2021covid} or predict user connections in a social one \cite{huang2020embedding}. 

In particular, \textit{link prediction} is a popular type of KGC which determines whether an edge exists between a given pair of nodes (\textit{e.g.}, whether two users of a social network might be acquainted in real life). Another variant, known as \textit{relation prediction}, aims to infer the existence of links of specific relation types (\textit{e.g.}, whether the two users are family, friends, or acquaintances) \cite{chen2020knowledge}. Link and relation predictions have been used for tasks such as drug repurposing for COVID-19 \cite{zhang2021drug}, polypharmacy side effect prediction \cite{carletti2021predicting, zitnik2018modeling}, and user recommendations in social networks \cite{daud2020applications}. Notably, while other questions may be answered through reasoning over KGs, such as subgraph or graph similarity (\textit{e.g.} representing and comparing chemical structures \cite{fang2021knowledge}), a majority of surveyed approaches utilize KGC.

\subsection{Rule-based Methods for KG Completion}
\label{subsection:rulebased_kg_completion}

\noindent Some of the simplest methods toward KGC utilize a set of rules that can be used for logical inference. The source of said rules varies between approaches. As previously stated in  \S\ref{subsection:introduction_to_logic}, \textit{ontologies} are formalizations of KB semantics and are often used to represent the unique patterns, relationships, and hierarchies within a specific knowledge domain \cite{dou2015semantic}. Notable examples in biomedicine, for example, include the Gene Ontology \cite{ashburner2000gene} and the Disease Ontology \cite{schriml2022human}. Therefore, ontologies based on description logics can serve as the source of such rules.

Since ontologies often comprise expert-curated knowledge \cite{dou2015semantic}, they can be especially beneficial if the application requires domain-specific knowledge \cite{kulmanov2021semantic}. However, ontologies do not exist for every specific domain, and existing ones may impose limitations on the type of predictions made. Alternatively, rules can be mined directly from the KG using ILP methods (discussed in  \S\ref{subsection:introduction_to_logic}) such as AMIE and its variants \cite{galarraga2013amie}. In these cases, rules are based on association patterns within the KG rather than the general application domain. Notably, Galárraga \etal \cite{galarraga2013amie} show that association rule mining can correspond to mining Horn clauses in sufficiently large KGs. Therefore, all approaches discussed here do \textit{logical} rule mining.

The major benefit of rule-based approaches is that they are inherently interpretable. One could refer back to the rules which governed the algorithm to get a human-readable understanding of how and why certain predictions were made. For example, SAFRAN \cite{ott2021safran}, which was inspired by AnyBURL \cite{meilicke2019anytime}, provides confidence-weighted, post-hoc explanations based on rules for every prediction. This allows model tuning, adjustment for incorrect predictions, explanations for end-users, and more \cite{ott2021safran}. Unfortunately, compared to other methodologies, such as those discussed in the next section, rule-based methods do not always achieve the same performance \cite{gema2023knowledge, meilicke2019anytime}. Although some types of logic, such as probabilistic and fuzzy logic (see \S\ref{subsection:introduction_to_logic}), allow the quantification of uncertainty, programs using them are still limited in their abilities to generalize beyond the patterns encoded in the rules \cite{gema2023knowledge}. Additionally, rule-based methods often suffer from scalability issues when faced with large KGs \cite{venugopal2014scaling, kazmi2017improving}, an increasingly pressing issue with the rise of big data and digitalization \cite{ji2021survey}.

\subsection{Graph Embedding-based Methods for KG Completion}
\label{subsection:embedding_kg_completion}

\noindent In contrast to rule-based methods, those which generate representative \textit{KG embeddings} (KGE) typically scale well to large datasets. A KGE  is a numerical vector representation of the KG constituents such that \textit{proximity} in the embedding space approximates some sort of \textit{similarity} in the original KG \cite{zhang2018link, li2020network}. Based on the graph notation defined in \mbox{\S\ref{subsection:intro_to_kgs}}, \(similarity(u, v) \approx proximity(e_u, e_v)\) for nodes \(u\) and \(v \in V\), in which \(e_{x}\) denotes the embedding for node \(x \in V\). From embeddings, one can re-construct parts of the KG to answer some predictive task, such as those mentioned in \S\ref{section:kg_complete}. Many popular KGE methods are based on NNs \cite{bordes2013translating, schlichtkrull2018modeling} and show competitive predictive performances in various fields \cite{zitnik2018modeling, ravindra2020disease}. In particular, GNNs are a popular category of methods to generate KGEs \cite{wu2020comprehensive}. GNNs account for connections between components; the input is typically a matrix representation of the graph and its constituents, such as node entities, edges and relation types, and node or edge features \cite{wu2020comprehensive}.

The way in which \textit{similarity} is defined and encoded is a major discrepancy between KGE methods \cite{li2020network}. For example, both DeepWalk \cite{perozzi2014deepwalk} and node2vec \cite{grover2016node2vec} assess the frequency in which nodes co-occur on random walks through the graph. In contrast, methods like TransE \cite{bordes2013translating} represent relationships as 1-1 translations in the embedding space. Notably, methodologies like RESCAL \cite{nickel2011three}, DistMult \cite{yang2014embedding}, and ComplEx \cite{trouillon2016complex} were groundbreaking for their stronger focus on relational information. Used alongside a \textit{link prediction} task (\mbox{\S\ref{section:kg_complete}}), for example, these approaches encode similar triples as similar elements in the embedding space. Thereafter, a \textit{decoder} function uses embeddings to compute \textit{scores}. For instance, the DistMult decoder \cite{yang2014embedding, schlichtkrull2018modeling}, which associates each relation \(r\) with a diagonal matrix \(R_r \in \mathbb{R}^{d\times d}\), computes the score of each triple with each relation, \(r(u, v)\) as:
\[
score_{r(u, v)} = e^{T}_{u}R_{r}e_{v}
\]
Models are trained to assign higher scores to positive triples in the KG and lower scores to negative ones that are not originally present in the data. Following training, the models are then used on some disjoint test set of triples. Negative triples which are scored highly are then regarded as novel positive predictions because they are considered to be similar to the positive samples in the KG \cite{yang2014embedding, trouillon2016complex, kong2019decompressing}. 

The Graph Convolutional Network (GCN) \cite{kipf2016semi} and its variants (\textit{e.g.}, Relational GCN (R-GCN) \cite{schlichtkrull2018modeling} and Graph Attention Network (GAT) \cite{velivckovic2017gat}) have expanded upon this concept with a \textit{message-passing paradigm}. These methods encode each node in the graph as a weighted combination of itself and its surrounding neighbors, thereby incorporating connectivity information. For example, the GCN uses the following equation, where \(e_{u}^{(0)}\) is the node's initial features:
\[
e_{u}^{(l+1)} = \sigma \left(W^{(l)}\sum_{v \in N(u) \cup \{u\}}\frac{e_{v}^{(l)}}{\sqrt{|N(v)| |N(u)|}} \right)
\]
\noindent The embedding for each node, \(u\), in the subsequent \textit{layer}, or iteration, \(l+1\), is computed by combining representations in the current layer, \(l\), from itself and all neighboring nodes, \(N(u)\). \(W^{(l)}\) is a learnable weight matrix, and \(\sigma\) is an activation function \cite{kipf2016semi, hamilton2020graph, daigavane2021understanding}. Thus, information is aggregated across edges at each hidden layer. However, this paradigm possesses several limitations as layers increase, including \textit{oversmoothing}, the convergence of node representations to the same value, making them less distinguishable \cite{wang2022guide, yan2022two, rusch2023survey}, and \textit{oversquashing}, the aggregation of too much information into a single vector, resulting in a less informative representation \cite{banerjee2022oversquashing, di2023does}. Furthermore, GNNs that utilize the message-passing paradigm tend to aggregate information between dissimilar node types (\textit{graph heterophily}) without accounting for the varied behaviors or relationships between them. This typically results in diminished or inconsistent performance \cite{yan2022two, cavallo2023gcnh}. While a number of promising extensions that capture the semantics of heterogeneous graphs have been studied recently \cite{shao2023heterogeneousgnn}, there are no strong guarantees that the encoding and aggregation mechanisms will be faithful to the heterogeneous relations in the graph.

Aside from these, KGE methods share other, general limitations. Unlike rule-based approaches, for example, the best performing KGE approaches tend to be black-box models, so no human-level explanations are generated along with predictions \cite{zhang2021survey}. Additionally, as many KGE methods utilize supervised learning, they require relatively large amounts of labeled data to capture relationships between entities \cite{zhao2022graph, li2023few}. Several neurosymbolic approaches help to overcome such limitations through characteristics discussed in the next section.


\section{Neurosymbolic Approaches for Reasoning over KGs}
\label{section:nesy_graphs}
\noindent As established, we observe a dichotomy between methods for reasoning on graph structures. Symbolic, rule-based methods (\S\ref{subsection:rulebased_kg_completion}) utilize domain knowledge and logic to infer new information in a naturally interpretable fashion. However, such approaches are limited in their performance and scalability to large KGs. In contrast, state-of-the-art approaches for learning KGEs (\S\ref{subsection:embedding_kg_completion}) tend to be black-box methods which aggregate information in domain-agnostic ways. As discussed in  \S\ref{subsection:neurosymbolic_ai}, recent studies in \emph{neurosymbolic AI} often combine aspects from deep learning and symbolic reasoning to mitigate their respective weaknesses and bridge such chasms \cite{tsamoura2021neural, garcez2020neurosymbolic, hitzler2022neuro}. Within this section, we survey several such approaches. Since each approach hybridizes the two fields in various ways, we aim to formalize and simplify the language we use by referring to each method's \textit{neural} and \textit{symbolic} modules. A \textit{neural} module refers to the use of a NN, often to produce KGEs, and a \textit{symbolic} module typically involves the use of logical rules. A list of the methods and representations each module could comprise, along with specific examples, is given in Table \ref{tab:modules}.

\begin{table*}[t!]
    \centering
    \begin{minipage}{.65\textwidth}
    \renewcommand{\arraystretch}{1.2}
    \begin{tabular}{p{2cm}p{3cm}p{5cm}}
         &  \textbf{Neural modules} & \textbf{Symbolic modules} \\
         \hline
         \textbf{Methods:} &  \textbf{KGE methods} (\textit{e.g.}, ComplEx \cite{trouillon2016complex}, R-GCN \cite{schlichtkrull2018modeling}) & \textbf{ILP / Logical rule mining} (\textit{e.g.}, AMIE \cite{galarraga2013amie}, SAFRAN \cite{ott2021safran}) \\
         
         &  \textbf{Other NN architectures} (\textit{e.g.,} RNN \cite{sherstinsky2020fundamentals}) & \textbf{Knowledge representation reasoners} (\textit{e.g.}, OWL reasoner \cite{mcguinness2004owl}) \\

         \hline
         \textbf{Representations:} &  \textbf{KGEs} &  \textbf{Ontology} (\textit{e.g.}, GO \cite{ashburner2000gene}, DO \cite{schriml2022human}) \\
         & \textbf{Logic embeddings}  &  \textbf{Logic program} (\textit{e.g.}, Horn clauses \cite{muggleton1994inductive}, MLN \cite{richardson2006markov})\\
         
    \end{tabular}
    \end{minipage}
    \begin{minipage}{.30\textwidth}
    \caption{Methodologies that could serve as neural and symbolic modules in a neurosymbolic approach for KGs. \textit{Methods} are the ways in which the \textit{Representations} may be obtained. Novel predictions can be made using one or more \textit{Representations}. Note that logic embeddings could be considered neural or symbolic.}
    \label{tab:modules}
    \end{minipage}
\end{table*}

Amongst the neurosymbolic approaches surveyed \cite{alshahrani2017neuro, chen2022qlogice, liu2021neural}, we note the following critical characteristics which are unique from symbolic or neural approaches alone:

\subsubsection{\textsc{Interpretability}} As mentioned, we often see a tradeoff between interpretability, which symbolic AI naturally possesses (in the form of verbal interpretability~\cite{tjoa2020survey}), and performance, in which black-box, KGE methods seem to dominate. In particular, enforcing interpretability often comes with a drop in predictive performance \cite{dziugaite2020enforcing, molnar2022} and might lead to infeasible computational complexity \cite{bertsimas2019price, dziugaite2020enforcing}. The surveyed neurosymbolic approaches foster interpretability and, in many cases, do not sacrifice performance heavily.\label{lab:interpretability}

\subsubsection{\textsc{Guided Training}} Some neurosymbolic approaches can integrate ontological or expert-defined knowledge into an otherwise data-driven approach \cite{alshahrani2017neuro, wickramarachchi2020evaluation}, bypassing the need for the model to learn known patterns. Furthermore, this guides learning toward more domain-congruous patterns. In some contexts, this is ideal for working with limited or small datasets. As described in \S\ref{subsection:embedding_kg_completion}, KGE methods tend to require a lot of labeled training data. Often, when a dataset is small, researchers devise relevant pre-training tasks on similar, larger datasets to steer the model's parameters toward a pertinent context \cite{li2016sparseness, yao2021understanding}. For example, one study pre-trained a GNN to predict atom-level perturbations in molecules to improve the model's overall ability to predict other molecular properties \cite{kiani2023utilizing}. However, pre-training is an additional and often computationally expensive training step. Neurosymbolic approaches may be a practical alternative.\label{lab:knowledge_guide}

\subsubsection{\textsc{Underrepresented Types}} Even if a labeled dataset is sufficiently large, sparsity or imbalance amongst labels may result in fewer instances of a certain type or subclass within that data. Many KGE approaches struggle to capture such underrepresented patterns \cite{zhao2022graph, li2023few}. Through KG augmentation, some neurosymbolic approaches pose novel ways to address this.\label{lab:low_data}

\subsubsection{\textsc{Heterogeneous Aggregation}} Heterogeneous KGs typically comprise nodes and edges of varying types, and these types may interact in different ways \cite{shao2023heterogeneousgnn}. As stated in \S\ref{subsection:embedding_kg_completion}, KGE methods using message-passing may perform inconsistently due to such heterophily. For example, biological KGs like Hetionet \cite{hetionet} comprise drugs, proteins, and diseases as nodes. While drug-protein edges may describe direct, physical interactions (\textit{e.g., binds}), drug-disease edges would constitute a more conceptual relationship (\textit{e.g., palliates}) \cite{hetionet}. Moreover, the features for such nodes would range from chemical structures to lettered sequences \cite{krix2023multigml}. The aggregation of features between dissimilar node types is a challenge which could be targeted via neurosymbolic methods: rather than adding complexity to KGE approaches to learn the relationships between node types, they can be represented through rules.\label{lab:heterogeneous}

\subsubsection{\textsc{Long-range Dependencies}} Due to oversquashing and oversmoothing (see \S\ref{subsection:embedding_kg_completion}), several GNN methods suffer from local receptive fields \cite{pasa2023empowering}, experiencing peak performances at two layers \cite{noor2023determining}. Consequently, they struggle to capture \mbox{\textit{long-range dependencies}} (\textit{i.e.}, relationships between nodes that are several \textit{hops} apart \cite{liu2021neural, banerjee2022oversquashing, di2023does}). Rule-based methods, though, can encode such relationships through a series of conjunctions in which each edge in the path is a binary predicate (\textit{e.g.,} \(\text{friends}(A, D) \leftarrow \text{likes}(A, B) \land \text{likes}(B, C) \land \text{likes}(C, D)\)), but inference scales poorly to large KGs. Specifically, \S\ref{subsubsection:path_based} describes a series of hybrid approaches which account for long-range dependencies.\label{lab:long_range}

\begin{figure*}[t!]
    \centering
    \begin{center}
        \input{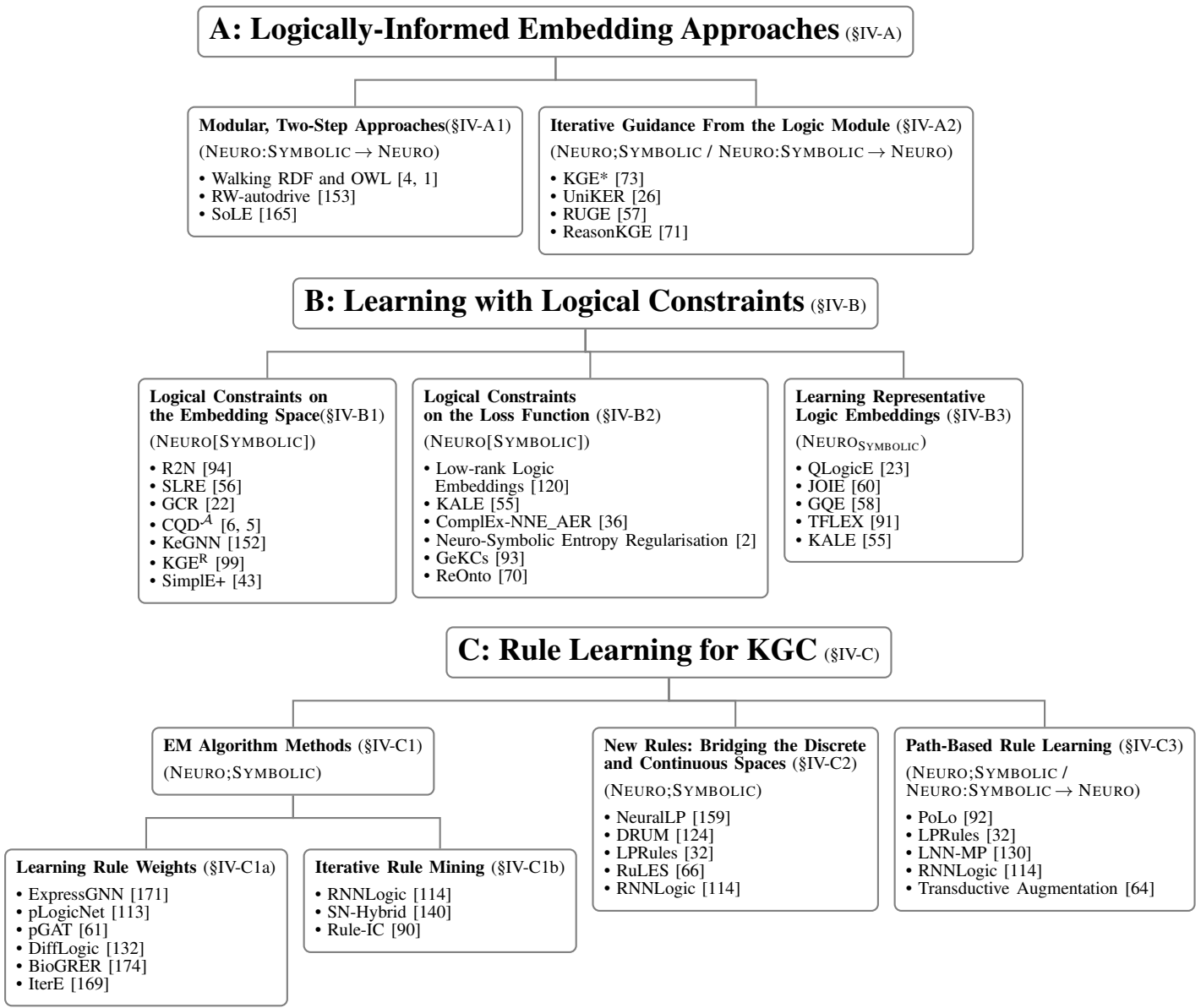}
        \begin{forest}
    mytree,
    [,phantom
        [\textbf{\Large{A}: Logically-Informed Embedding Approaches}  (\S\ref{subsection:fusion})
            [{\textbf{Modular, Two-Step Approaches}(\S\ref{subsection:fusion1})\\[0.5em]
            (\textsc{Neuro:Symbolic $\rightarrow$ Neuro})\\[0.5em]
                \textbullet \ Walking RDF and OWL \cite{alshahrani2017neuro, agibetov2018fast}\\
                \textbullet \ RW-autodrive \cite{wickramarachchi2020evaluation}\\
                \textbullet \ SoLE \cite{zhang2019enhanced}
            }]
            [{\textbf{Iterative Guidance From the Logic Module}    (\S\ref{subsection:fusion2})\\[0.5em]
            (\textsc{Neuro;Symbolic} / \textsc{Neuro:Symbolic $\rightarrow$ Neuro})\\[0.5em]
                \textbullet \ KGE* \cite{kaoudi2022towards}\\
                \textbullet \ UniKER \cite{cheng2020uniker}\\
                \textbullet \ RUGE \cite{guo2018knowledge}\\
                \textbullet \ ReasonKGE \cite{jain2021improving}
            }
            ]
        ]
    ]
\end{forest}\\[1em]
        \begin{forest}
    mytree,
    [,phantom
        [\textbf{\Large{B}: Learning with Logical Constraints}  (\S\ref{subsection:logical_constraints})
            [{\textbf{Logical Constraints on}\\\textbf{the Embedding Space}(\S\ref{logcon_on_embed})\\[0.5em]
            (\textsc{Neuro[Symbolic]})\\[0.5em]
                \textbullet \ R2N \cite{marra2021learning}\\
                \textbullet \ SLRE \cite{guo2020knowledge}\\
                \textbullet \ GCR \cite{HChenGCR}\\
                \textbullet \ CQD\(\mathbf{^{\mathcal{A}}}\) \cite{cqd, cqd_a}\\
                \textbullet \ KeGNN \cite{werner2023knowledge}\\
                \textbullet \ KGE\textsuperscript{R} \cite{minervini2017regularizing}\\
                \textbullet \ SimplE+ \cite{fatemi2019improved}
            }]
            [{\textbf{Logical Constraints}\\\textbf{on the Loss Function} (\S\ref{rule-based-restrictions-on-possible-predictions})\\[0.5em]
            (\textsc{Neuro[Symbolic]})\\[0.5em]
                \textbullet \ Low-rank Logic \\ \; Embeddings \cite{rocktaschel2015injecting}\\
                \textbullet \ KALE \cite{Guo2016JointlyEK}\\
                \textbullet \ ComplEx-NNE_AER \cite{ding2018improving}\\
                \textbullet \ Neuro-Symbolic Entropy Regularisation \cite{ahmed2022neuro}\\
                \textbullet \ GeKCs \cite{loconte2023turn}\\
                \textbullet \ ReOnto \cite{jain2023reonto}
            }]
            [{\textbf{Learning Representative}\\\textbf{Logic Embeddings} (\S\ref{logcon_log_embed})\\[0.5em]
            (\textsc{Neuro}\textsubscript{\textsc{Symbolic}})\\[0.5em]
                \textbullet \ QLogicE \cite{chen2022qlogice}\\
                \textbullet \ JOIE \cite{hao2019universal}\\
                \textbullet \ GQE \cite{hamilton2018embedding} \\
                \textbullet \ TFLEX \cite{lin2024tflex}\\
                \textbullet \ KALE \cite{Guo2016JointlyEK}
            }]
        ]
    ]
\end{forest}\\[1em]
\begin{forest}
    mytree,
    [,phantom
        [\textbf{\Large{C}: Rule Learning for KGC} (\S\ref{subsection:learning_rules})
            [{\textbf{EM Algorithm Methods} (\S\ref{subsection:iterative})\\[0.5em]
            (\textsc{Neuro;Symbolic}) 
                },l sep+=2pt,
                [{\textbf{Learning Rule Weights} (\S\ref{learning_rule_weights})\\[0.5em]
                    \textbullet \ ExpressGNN \cite{zhang2020efficient}\\
                    \textbullet \ pLogicNet \cite{qu2019probabilistic}\\
                    \textbullet \ pGAT \cite{harsha2020probabilistic}\\
                    \textbullet \ DiffLogic \cite{shengyuan2023differentiable}\\
                    \textbullet \ BioGRER \cite{zhao2020biomedical} \\
                    \textbullet \ IterE \cite{zhang2019iteratively}
                },
                edge path={
                  \noexpand\path [draw, \forestoption{edge}] (!u.parent anchor) -- +(0,-10pt) -| (.child anchor)\forestoption{edge label};
                }]
                [{\textbf{Iterative Rule Mining} (\S\ref{iterative_rule_mining})\\[0.5em]
                    \textbullet \ RNNLogic \cite{qu2020rnnlogic} \\
                    \textbullet \ SN-Hybrid \cite{suresh2020hybrid} \\ 
                    \textbullet \  Rule-IC \cite{lin2021rule} 
                },
                edge path={
                  \noexpand\path [draw, \forestoption{edge}] (!u.parent anchor) -- +(0,-10pt) -| (.child anchor)\forestoption{edge label};
                }]
            ]
            [{\textbf{New Rules: Bridging the Discrete}\\\textbf{and Continuous Spaces} (\S\ref{subsection:rulebased})\\[0.5em]
            (\textsc{Neuro;Symbolic})\\[0.5em]
                \textbullet \ NeuralLP \cite{yang2017differentiable}\\ 
                \textbullet \ DRUM \cite{sadeghian2019drum}\\ 
                \textbullet \ LPRules \cite{dash2021lprules}\\
                \textbullet \ RuLES \cite{ho2018rule} \\
                \textbullet \ RNNLogic \cite{qu2020rnnlogic}
            }]
            [{\textbf{Path-Based Rule Learning} (\S\ref{subsubsection:path_based})\\[0.5em]
            (\textsc{Neuro;Symbolic} /\\
            \textsc{Neuro:Symbolic $\rightarrow$ Neuro})\\[0.5em]
                \textbullet \ PoLo \cite{liu2021neural}\\
                \textbullet \ LPRules \cite{dash2021lprules}\\
                \textbullet \ LNN-MP \cite{sen2021combining}\\
                \textbullet \ RNNLogic \cite{qu2020rnnlogic}\\
                \textbullet \ Transductive Augmentation \cite{hirose2021transductive}
            }]
        ]
    ]
\end{forest}
    \end{center}
    \caption{Taxonomy of neurosymbolic approaches for graph reasoning.}
    \label{fig:taxonomy}
\end{figure*}

Within the remaining sections, we discuss various approaches along with their strengths and weaknesses, referring specifically to these enumerated characteristics as guides for discussion. Importantly, we group the approaches according to the taxonomy in Figure \ref{fig:taxonomy} to facilitate easy comparison. It is independent of Kautz's classification, introduced in \S\ref{subsection:neurosymbolic_ai}, as we aim for a finer-grained division, specialized to KGs. However, there is naturally some overlap, so we indicate which types, according to Kautz, are present in each of our classes. Note that Kautz's types 1 and 2 do not appear in our taxonomy, as none of the surveyed approaches align with them. Additionally, we summarize the main points of this survey within Table~\ref{tab:summary_table} and provide a curated a repository of available code for each work on GitHub\footnote{\url{https://github.com/NeSymGraphs}}.

\subsection{Logically-Informed Embedding Approaches}
\label{subsection:fusion}

\noindent We begin by introducing some of the most intuitive neurosymbolic approaches for reasoning on graph structures. This section is depicted by Fig. \ref{fig:taxonomy}-A, and the recently discussed characteristics on \mbox{\refguide} and \mbox{\refunderep} are most prominent. To combine the benefits of symbolic and neural modules, these approaches modularize the two and then feed the results from the former into the latter. Since symbolic approaches are often based on expert-defined rules, they can be viewed as methods to \textit{extend} the ground truth. Therefore, inference by the symbolic module is used as a preliminary, \textit{KG augmentation} step, which feeds into the neural module (typically a KGE method) for further processing and prediction. In some situations, this KG augmentation step could be exploited to expand small datasets or ameliorate dataset imbalance, thereby harnessing the \mbox{\refunderep} characteristic. Because the symbolic knowledge is integrated into the training set, these approaches are much like Kautz's \mbox{\textsc{Neuro:Symbolic~$\rightarrow$~Neuro}} category. We divide these approaches into two subcategories:

\subsubsection{\textbf{Modular, Two-Step Approaches}}
\label{subsection:fusion1}

\begin{figure*}[t!]
    \centering
    \input{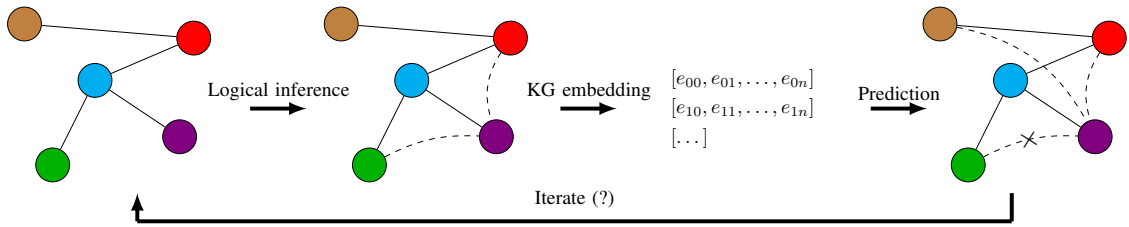}
    \scalebox{\diagramscale}{\def\colorA{brown}
\def\colorB{red}
\def\colorC{cyan}
\def\colorD{black!30!green}
\def\colorE{violet}
\def\diagramscale{0.75}
\begin{tikzpicture}
    \tikzset{
    node_style/.style={shape=circle, draw=black, minimum size=6mm}
    }
    
    \node[node_style, fill=\colorA] (A1) at (0, 0) {};
    \node[node_style, fill=\colorB] (B1) at (3, -0.25) {};
    \node[node_style, fill=\colorC] (C1) at (1.25, -1) {};
    \node[node_style, fill=\colorD] (D1) at (0.5, -2.5) {};
    \node[node_style, fill=\colorE] (E1) at (2.75, -2) {};

    \path[-] (A1) edge node[] {} (B1);
    \path[-] (B1) edge node[] {} (C1);
    \path[-] (C1) edge node[] {} (D1);
    \path[-] (C1) edge node[] {} (E1);

    \node[node_style, fill=\colorA, right = 5cm of A1] (A2) {};
    \node[node_style, fill=\colorB, right = 5cm of B1] (B2) {};
    \node[node_style, fill=\colorC, right = 5cm of C1] (C2) {};
    \node[node_style, fill=\colorD, right = 5cm of D1] (D2) {};
    \node[node_style, fill=\colorE, minimum size=6mm, right = 5cm of E1] (E2) {};

    \path[-] (A2) edge node[] {} (B2);
    \path[-] (B2) edge node[] {} (C2);
    \path[-] (C2) edge node[] {} (D2);
    \path[-] (C2) edge node[] {} (E2);
    \path[-] (B2) edge[dashed, bend right=25] node[] {} (E2);
    \path[-] (D2) edge[dashed, bend left=15] node[] {} (E2);

    \node[below right = 1.25cm and 5.5cm of A2, text centered, text height=0cm] (O) {
        $\begin{aligned}
             &[e_{00}, e_{01}, \dots, e_{0n}] \\
             &[e_{10}, e_{11}, \dots, e_{1n}] \\
             &[\dots ]
        \end{aligned}$
    };

    \node[node_style, fill=\colorA, right = 10cm of A2] (A3) {};
    \node[node_style, fill=\colorB, right = 10cm of B2] (B3) {};
    \node[node_style, fill=\colorC, right = 10cm of C2] (C3) {};
    \node[node_style, fill=\colorD, right = 10cm of D2] (D3) {};
    \node[node_style, fill=\colorE, minimum size=6mm, right = 10cm of E2] (E3) {};

    \path[-] (A3) edge node[] {} (B3);
    \path[-] (B3) edge node[] {} (C3);
    \path[-] (C3) edge node[] {} (D3);
    \path[-] (C3) edge node[] {} (E3);
    \path[-] (A3) edge[dashed, bend left=25] node[] {} (E3);
    \path[-] (B3) edge[dashed, bend right=25] node[] {} (E3);
    \path[-] (D3) edge[dashed, bend left=15] node[sloped, anchor=center] {\Large $\times$} (E3);

    \path[-latex, line width=2pt] (4, -1.5) edge node[above] {Logical inference} (5, -1.5);
    \path[-latex, line width=2pt] (9.5, -1.5) edge node[above] {KG embedding} (10.5, -1.5);
    \path[-latex, line width=2pt] (15, -1.5) edge node[above] {Prediction} (16, -1.5);
    \path[-, line width=2pt] (17.5, -3) edge (17.5, -3.5);
    \path[-, line width=2pt] (17.5, -3.5) edge node[sloped, anchor=center, above=1mm] {Iterate (?)} (2, -3.5);
    \path[-latex, line width=2pt] (2, -3.5) edge (2, -3);
\end{tikzpicture}}
    \caption{\textbf{Logically-Informed Embedding Approaches} (\S\ref{subsection:fusion1}) These approaches augment the KG with logical inference, then use a KGE method on the augmented KG. As shown by the lowermost arrow, some approaches iterate between logical inference and KGE predictions, using the latter as ground truth for the next iteration's input (\S\ref{subsection:fusion2}).}
    \label{fig:modular_two_step}
\end{figure*}

This is depicted by the leftmost subcategory of Fig. \ref{fig:taxonomy}-A and demonstrates the first example of the \mbox{\refunderep} characteristic. These approaches involve a simple, unidirectional flow of information from the symbolic module to the neural module. For example, Alshahrani \etal \cite{alshahrani2017neuro} developed a biological KG from various sources and employed an ontological reasoner to augment and control the scope of the graph. With the reasoner, a fully deduced graph is thus obtained with newly inferred edges. As illustrated in Fig. \ref{fig:modular_two_step}, embeddings are subsequently generated from this augmented KG using the DeepWalk algorithm \cite{perozzi2014deepwalk} (see \S\ref{subsection:embedding_kg_completion}). Finally, link prediction is performed using embeddings as input. They tested their pipeline, which they coined \textbf{Walking RDF and OWL}, on a drug repurposing task to determine whether two drugs share indications, represented by edges between drug and disease nodes. They found that prediction performance improved with the augmented KG. Agibetov and Samwald \cite{agibetov2018fast} aimed to improve \textbf{Walking RDF and OWL} with an alternative, log-linear embedding method in the neural module, showing that different combinations of symbolic and neural modules may affect predictions. Consequently, one may wonder how consistently the KG augmentation step improves performance.

Another study by Wickramarachchi \etal \cite{wickramarachchi2020evaluation}, which we will refer to as \textbf{RW-autodrive}, explored this, using varying neural modules in an autonomous driving application. For this study's scene understanding task, data was derived from sensors, such as LIDAR and RADAR, as well as video data taken from the car position. They compared the performances of three different KGE algorithms between both raw KGs and KGs augmented by an ontological reasoner. They showed that, regardless of the KGE algorithm used, better performance was achieved on the augmented KGs, suggesting that incorporation of domain knowledge into KGE methods is especially useful within the autonomous driving domain \cite{wickramarachchi2020evaluation}.


While incorporating domain-specific knowledge often improves model performance, the obvious limitation is the requirement for hard rules in the form of an ontology or expert-defined rule set. Thus, there is no opportunity to represent uncertainty or consider new rules. In contrast, \textbf{Soft Logical Rules Enhanced Embedding (SoLE)} \cite{zhang2019enhanced} utilizes soft rules, as discussed in \S\ref{subsection:introduction_to_logic}, alongside corresponding confidences learnt from KGs. Essentially, \textbf{SoLE}'s symbolic module acts as a \textit{rule engine} by mining soft rules from the KG. In the KG augmentation step, new groundings, or triples, are inferred by those soft rules and subsequently combined with existent KG triples. However, \textbf{SoLE} does not scale well to larger KGs due to the rule-mining process; an additional rule-pruning step would be useful, a tactic discussed in \S\ref{subsection:rulebased_kg_completion}.

Overall, two-step approaches are clear and straightforward modular structures to compute KGEs, with room for flexibility on both the symbolic and neural sides. However, their usefulness depends upon the availability of relevant rule sets and face scalability issues when rule-mining is involved. Furthermore, these approaches do not take full advantage of the capabilities of either the symbolic or neural modules. In particular, as noted in \S\ref{section:nesy_graphs}, many neurosymbolic methods are specifically designed to promote interpretability. However, these two-step methods have the same black-box limitation in their neural modules as any other KGE or GNN approach might since the logic module is only used for KG augmentation. 

\subsubsection{\textbf{Iterative Guidance from the Logic Module}}
\label{subsection:fusion2}

A few approaches extend the two-step pattern to increase interaction between the modules. We classify these approaches into a second subcategory, depicted in the rightmost subsection of Fig. \ref{fig:taxonomy}-A and characterized by \mbox{\refguide}. In the first subcategory, the directionality of information is one-way: logical inference from the symbolic module informs the KGE method, but the reverse is not true; the output of the embedding method is not used to alter the logical inference step. The minimal interaction between the symbolic and neural modules could, therefore, limit the practicality of these methods. In practice, utilizing the output of logical inference as ground truth labels for the neural module could be fallacious as it assumes that logical inference is monotonic, \textit{i.e.}\ it always yield true results. In reality, expert-defined rules often have exceptions \cite{garcez2020neurosymbolic, furman2020exceptions}, so regulating the symbolic module could be useful in some cases.

In light of the above, Kaoudi \etal\!'s \textbf{KGE*} \cite{kaoudi2022towards}, Cheng \etal\!'s \textbf{Unified Framework for Knowledge Graph Inference (UniKER)} \cite{cheng2020uniker}, and Guo \etal\!'s \textbf{Rule-Guided Embedding (RUGE)} \cite{guo2018knowledge} all follow similar patterns: the two modules inform one another in an iterative style, as shown in Fig. \ref{fig:modular_two_step}. Both \textbf{KGE*} and \textbf{UniKER} use Horn rules and forward chaining, a form of logical inference, to augment the KG as in the previous section. Alternatively to Horn rules, \textbf{KGE*} also supports using an ontology for the inference step, while \textbf{RUGE} uses soft rules. After KG augmentation in the respective symbolic module, a KGE method is trained, and the resulting predictions are used to refine the KG by eliminating the least probable edges and adding the most probable ones. Thereafter, the newly refined KG is passed back to the beginning of the pipeline for the next iteration. One shared shortcoming of the these approaches, however, is that they are designed mainly to generate positive predictions; this can lead to an increased number of false positive predictions. As a solution, another iterative approach, called \textbf{ReasonKGE} \cite{jain2021improving}, adds a step in which negative samples are also updated.

By using an iterative process, these methods introduce bidirectionality in which both symbolic and neural modules inform one another by taking turns to refine the KG. One might argue, then, that these approaches fit most closely with Kautz's third category of \mbox{\textsc{Neuro;Symbolic}} AI, rather than the fourth. However, there is a key caveat: the parameters of the symbolic module are \textit{not} updated as in the neural module; the rules of the symbolic module are static. Therefore, we still view these approaches as primarily neural. Methods which implement dynamic symbolic modules are discussed later within \S\ref{subsection:iterative}. Ultimately, we consider the static symbolic modules in these approaches as simply \textit{guiding} model training rather than \textit{driving} it. Nevertheless, guidance via the symbolic module allows these approaches to fulfill the \mbox{\refguide} characteristic. For example, if a user chooses to implement domain-specific rules, such as that from an ontology, into the symbolic module, then groundings from such rules will be included as input into the neural module. Therefore, the neural module will learn to encode patterns consistent with such rules. These approaches move conceptually closer, then, to those which apply logical constraints onto neural modules, the topic of the next section.

\subsection{Learning with Logical Constraints}
\label{subsection:logical_constraints} 

\noindent In contrast to using two separate modules, a different neurosymbolic pattern on KGs involves imposing the symbolic module, in the form of logical or rule-based constraints, onto the neural module. This pattern corresponds to Fig. \ref{fig:taxonomy}-B. In this case, the focus is on training the neural module, but rules are used both to incorporate domain-specific knowledge and to limit the scope of predictions possible. These approaches are, therefore, the epitome of the \mbox{\refguide} characteristic, as their primary goals include guiding neural training via the symbolic module. For reasons discussed toward the end of  this section, these approaches tend to be the weakest with regard to \mbox{\refinterp}. However, some are particularly useful for integrating heterogeneous information, \textit{i.e.} the 
\mbox{\refhetagg} characteristic. This section fits most closely with Kautz's sixth category of \mbox{\textsc{Neuro[Symbolic]}} AI because the constraints often act as logical checkpoints within the neural engines.

These approaches fundamentally differ based on where logical constraints are imposed in the learning pipeline. Some approaches, discussed in \S\ref{logcon_on_embed} below, impose them directly onto the learned embeddings, while other approaches, discussed in \S\ref{rule-based-restrictions-on-possible-predictions}, influence model training by restricting the predictions made from embeddings. \S\ref{logcon_log_embed}, for its part, discusses methods that learn entirely separate embeddings for logical constraints.

\subsubsection{\textbf{Logical Constraints on the Embedding Space}}
\label{logcon_on_embed}

\begin{figure*}[b!]
    \centering
    \input{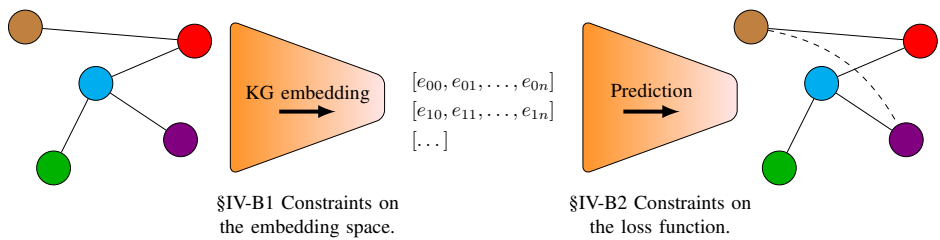}
    \scalebox{\diagramscale}{\def\colorA{brown}
\def\colorB{red}
\def\colorC{cyan}
\def\colorD{black!30!green}
\def\colorE{violet}
\def\diagramscale{0.75}
\begin{tikzpicture}
    \tikzset{
        node_style/.style={
            shape=circle, 
            draw=black, 
            minimum size=6mm
        },
        myfilter/.style={%
            draw,
            trapezium,
            shape border rotate=270,
            text width=2.5cm,
            text height=0.5cm,
            align=center,
            trapezium angle=70,
            shading=axis,
            shading angle=90,
            left color=orange!85,
            right color=red!10,
            opacity=0.4,
            rounded corners
        }
    }
    
    \node[node_style, fill=\colorA] (A1) at (0, 0) {};
    \node[node_style, fill=\colorB] (B1) at (3, -0.25) {};
    \node[node_style, fill=\colorC] (C1) at (1.25, -1) {};
    \node[node_style, fill=\colorD] (D1) at (0.5, -2.5) {};
    \node[node_style, fill=\colorE] (E1) at (2.75, -2) {};

    \path[-] (A1) edge node[] {} (B1);
    \path[-] (B1) edge node[] {} (C1);
    \path[-] (C1) edge node[] {} (D1);
    \path[-] (C1) edge node[] {} (E1);

    \node[below right = 1.25cm and 6.5cm of A1, text centered, text height=0cm] (O) {
        $\begin{aligned}
             &[e_{00}, e_{01}, \dots, e_{0n}] \\
             &[e_{10}, e_{11}, \dots, e_{1n}] \\
             &[\dots ]
        \end{aligned}$
    };

    \node[node_style, fill=\colorA, right = 12.25cm of A1] (A2) {};
    \node[node_style, fill=\colorB, right = 12.25cm of B1] (B2) {};
    \node[node_style, fill=\colorC, right = 12.25cm of C1] (C2) {};
    \node[node_style, fill=\colorD, right = 12.25cm of D1] (D2) {};
    \node[node_style, fill=\colorE, minimum size=6mm, right = 12.25cm of E1] (E2) {};

    \path[-] (A2) edge node[] {} (B2);
    \path[-] (B2) edge node[] {} (C2);
    \path[-] (C2) edge node[] {} (D2);
    \path[-] (C2) edge node[] {} (E2);
    \path[-] (A2) edge[dashed, bend left=25] node[] {} (E2);

    \node[myfilter] (f1) at (5, -1.25) {};
    \node[below = 0.75cm of f1, align=center] (f1txt) {\S\ref{logcon_on_embed} Constraints on\\the embedding space.};
    \node[myfilter] (f2) at (11.25, -1.25) {};
    \node[below = 0.75cm of f2, align=center] (f2txt) {\S\ref{rule-based-restrictions-on-possible-predictions} Constraints on\\the loss function.};
    
    \path[-latex, line width=2pt] (4.5, -1.5) edge node[above] {KG embedding} (5.5, -1.5);
    \path[-latex, line width=2pt] (10.6, -1.5) edge node[above=1mm] {Prediction} (11.6, -1.5);
    
\end{tikzpicture}}
    \caption{\textbf{Learning with Logical Constraints}. (\S\ref{logcon_on_embed}) Logical constraints on the embedding space (first filter). These methods drive training by imposing logical constraints onto the embedding space, such as through a transformation. (\S\ref{rule-based-restrictions-on-possible-predictions}) Alternatively, logical constraints on the loss function (second filter). These methods drive which predictions are made by encoding logical constraints into the loss function, such as with a penalty term.}
    \label{fig:constraints_emb_space_or_loss_func}
\end{figure*}



One way to approach imposing logical constraints onto KGE methods is to alter the embedding space in some meaningful way, as depicted in Fig. \ref{fig:constraints_emb_space_or_loss_func} and the leftmost branch of the taxonomy in Fig. \ref{fig:taxonomy}-B. Such approaches are also particularly useful for the \mbox{\refguide} and \mbox{\refhetagg} characteristics. The \textbf{Relational Reasoning Network (R2N)} \cite{marra2021learning}, for example, uses structural relational information as constraints on GNN training, sharing quite similar goals to the previously mentioned R-GCN \cite{schlichtkrull2018modeling} (\S\ref{subsection:embedding_kg_completion}). Using the phrase ``relational reasoning in the latent space'' to describe their process, they transform the latent space embeddings in a way that is dependent upon each node’s neighbors in the original space. They argue that the transformation produces KGEs encoding information about both the individual nodes and the surrounding graph topology. In a similar manner to \textbf{R2N}, \textbf{Soft Logical Regularity in Embeddings (SLRE)} \cite{guo2020knowledge} uses soft logical rules as constraints on \textit{relation} embeddings. These rules are fused into the KGE process by imposing rule-based regularization onto generated embeddings. Consequently, embeddings are potentially more useful for generating predictions aligned with domain-specific or logic-based paradigms.

The emphasis on using relational information seen in \textbf{R2N} and \textbf{SLRE} is also key to a study by H. Chen \etal \cite{HChenGCR}, who propose \textbf{Graph Collaborative Reasoning (GCR)}. The \textbf{GCR} extends the vanilla GNN to encode whole triples, rather than nodes alone, focusing their analysis on adjacent edges. Conceptually, they transform the KG structure into a series of logical expressions which are used to predict the probability that a novel edge is implied by its adjacent edges. This re-frames the link prediction problem as a neural logic reasoning one. More specifically, they accomplish this by applying logical regularizers to MLPs to simulate logical operations; these MLPs are then applied to triple embeddings to generate predictions. Furthermore, to avoid the requirement of specifying rules by hand, they introduce a method to learn potential rules, formulated as Horn clauses, from the available training triples; this benefit foreshadows an approach presented in \S\ref{subsection:learning_rules}. Additionally, the authors claim that their method is more scalable to large, complex KGs than other neurosymbolic methods. However, we note that the scope of applications on which their method might be applicable could be limited. The logical expressions by which they perform link prediction rely on the idea that closely connected neighbors likely share similar interaction patterns. While this is often true for social KGs, it may not suit other domains in which entities with similar patterns might avoid interacting closely.

Two additional studies, \textbf{Knowledge Enhanced Graph Neural Networks (KeGNN)} \cite{werner2023knowledge} and \textbf{Continuous Query Decomposition (CQD)} \cite{cqd} also view KGs in terms of logical expressions. Both demonstrate that, after using KGE methods for simple link prediction, $t$-norms (\mbox{\S\ref{subsection:introduction_to_logic}}) and $t$-conorms (for disjunctions) can be applied to embeddings to answer more complex queries comprising \textit{multiple} links. In theory, this approach could be useful for \mbox{\reflongr} if the queries involve a series of conjunctions; this idea is elaborated upon in \mbox{\S\ref{subsubsection:path_based}}. In particular, \textbf{CQD} demonstrates its usefulness for queries of up to eight links. Its extension, \textbf{CQD}\(\mathbf{^{\mathcal{A}}}\) \cite{cqd_a}, imposes learnable adaptation functions to accommodate for the interactions between parts of a complex query. Such adaptation functions could be a way to cope with graph heterophily, mentioned in \mbox{\S\ref{subsection:embedding_kg_completion}}.

Minervini \etal also explore the use of logical constraints as regularizers with an approach we will refer to as \textbf{KGE\textsuperscript{R}} \cite{minervini2017regularizing}. Specifically, they apply constraints to encode information about relations that may mean the same thing, such as \textit{PartOf} and \textit{ComponentOf}, through a model-dependent transformation on the embeddings. They derive these transformations for three different KGE methods and discover that they all yield more accurate link predictions with regularization. As with \textbf{SLRE}, applying constraints directly to the embeddings was found to be effective in integrating domain knowledge without impairing scalability \cite{minervini2017regularizing}. However, unlike \textbf{SLRE}, \textbf{KGE\textsuperscript{R}} is not model agnostic, requiring the user to derive a useful transformation for each individual KGE method. We note that this also poses the risk that the regularization method might not generalize well to any given embedding method.

Instead of model-dependent regularization, both \textbf{SimplE+} \cite{fatemi2019improved} and \textbf{SLRE} enforce constraints on embeddings by requiring non-negativity. The authors of \textbf{SimplE+} explain that their motivation in using non-negativity, specifically, is to enforce the \textit{subsumption} axiom from OWL's semantics \cite{mcguinness2004owl}. Essentially, subsumption determines whether something is a subclass or subproperty of another. By enforcing that generated entity embeddings must be non-negative, they prove that subsumption is enforced. Furthermore, they also impose a constraint on relations which can be subsumed from one another. Their method outperformed simple logical inference as well as the basic embedding method without non-negativity constraints \cite{fatemi2019improved}.

Imposing logical constraints onto the embedding space is ideal when one wishes to encode specific information about the relational nature of the graph. In particular, incorporating relational or ontological relationships into KGEs could be considered as a solution to the challenges posed by heterogeneous KGs, as described in \S\ref{section:nesy_graphs}. These neurosymbolic approaches take into account the unique and varied relationships between different node types when generating predictions, thus harnessing the \mbox{\refhetagg} characteristic. However, depending upon the needs of the researcher, such constraints may be too restrictive. Instead of applying constraints onto the embedding space, thereby limiting the types of patterns encoded, another subcategory of approaches, discussed in the next section, applies constraints to the processes of decoding the embeddings and generating predictions.

\subsubsection{\textbf{Logical Constraints on the Loss Function}}
\label{rule-based-restrictions-on-possible-predictions}

A drawback of popular KGE algorithms (namely RESCAL, TransE, HolE, and ComplEx, as discussed in \S\ref{subsection:fusion}) is that they only implicitly learn higher-order relationships among triples in the KG, requiring more data to learn its latent logical structure. Moreover, this learned structure is confined to the ground triples available during training. These limitations can be overcome by the next subcategory of approaches, depicted by the centre branch of Fig. \ref{fig:taxonomy}-B. These use logical constraints to guide training toward predictions which align with some knowledge base or logical entailment (Fig. \ref{fig:constraints_emb_space_or_loss_func}), aligning with the \mbox{\refguide} characteristic. Here, this is accomplished by altering the loss function.

For example, \textbf{ReOnto} \cite{jain2023reonto} integrates ontological axioms as a bias term into the loss function of a GNN, coercing the training process to adhere to biomedical knowledge. Despite reporting competitive performance, \textbf{ReOnto} requires a domain-specific ontology. Alternatively, an approach called \textbf{Low-rank Logic Embeddings (LRLE)} \cite{rocktaschel2015injecting} integrates logical constraints as differentiable loss functions to be jointly optimized. Given logical rules of the form
\[
    \text{relation\_a}(X,Y) \rightarrow \text{relation\_b}(X,Y),
\]
two distinct processes are carried out. First, the rules are applied to the existing training triples to generate new ones, as described in \S\ref{subsection:fusion}. Additionally, this encourages the learned embeddings to encode inter-relation structures. Second, a unified set of differentiable loss functions corresponding to both the set of training triples and the logical rules is created using fuzzy $t$-norms (see \S\ref{subsection:introduction_to_logic}). This creates a joint optimization problem over capturing factual information while obeying the given constraints. A summation of the individual losses is formulated as a log-likelihood loss so that embeddings which assign a high marginal probability to rule and triple satisfaction are preferred. The authors argue that the framing of constraint satisfaction as a probabilistic model allows the embeddings to be unaffected by noisy data.

However, \textbf{LRLE} is also limited in that it only models existent relations and entity-pairs. Consequently, rules will not be discovered for entity-pairs that do not appear in the training data. This shortcoming is significant as most KGs are inherently incomplete. This drawback is tackled by Guo \etal\!'s method, \textbf{Embeddings by jointly modeling Knowledge And Logic (KALE)} \cite{Guo2016JointlyEK}, which performs the same unification of triples and rules, but also explicitly models entities on their own. This allows for novel relations between entities to be predicted, as demonstrated by a recent study \cite{spillo2022knowledge} which employed \textbf{KALE} as an effective recommender system. \textbf{KALE} extends TransE and maintains the same competitive time and space complexity thereof. Moreover, while their approach uses a pairwise ranking loss, the authors note that the generality of their approach facilitates the use of different losses such as with \textbf{LRLE}.

\begin{figure*}[b!]
    \centering
    \input{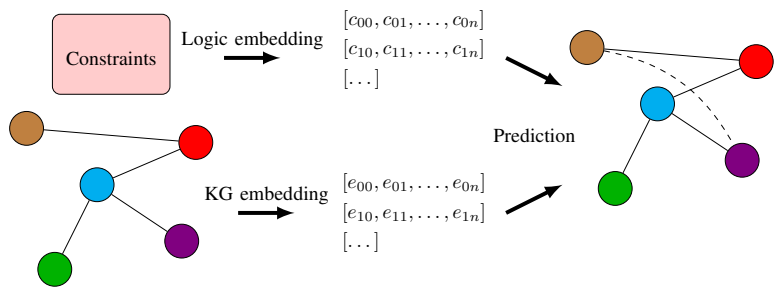}
    \scalebox{\diagramscale}{\begin{tikzpicture}
    \tikzset{node_style/.style={shape=circle, draw=black, minimum size=6mm}}
    
    \node[node_style, fill=\colorA] (A1) at (0, 0) {};
    \node[node_style, fill=\colorB] (B1) at (3, -0.25) {};
    \node[node_style, fill=\colorC] (C1) at (1.25, -1) {};
    \node[node_style, fill=\colorD] (D1) at (0.5, -2.5) {};
    \node[node_style, fill=\colorE] (E1) at (2.75, -2) {};

    \path[-] (A1) edge node[] {} (B1);
    \path[-] (B1) edge node[] {} (C1);
    \path[-] (C1) edge node[] {} (D1);
    \path[-] (C1) edge node[] {} (E1);

    \node[below right = 1.25cm and 5.25cm of A1, text centered, text height=0cm] (Emb) {
        $\begin{aligned}
             &[e_{00}, e_{01}, \dots, e_{0n}] \\
             &[e_{10}, e_{11}, \dots, e_{1n}] \\
             &[\dots ]
        \end{aligned}$
    };

    \path[-latex, line width=2pt] (3.75, -1.5) edge node[above] {KG embedding} (4.75, -1.5);

    \node[text centered] (Constraints) at (1.5, 1.25) {Constraints};
    \node[] (ghost) at (2, 1.5) {};
    \node[draw=black, fill=red!20, opacity=0.2, rounded corners, fit=(Constraints) (ghost), inner ysep=0.4cm] (constraintsbox) {};
    \node[text centered] at (1.5, 1.25) {Constraints};
    
    \path[-latex, line width=2pt] (3.5, 1.25) edge node[above] {Logic embedding} (4.5, 1.25);

    \node[above = 2cm of Emb, text centered, text height=0cm] (CEmb) {
        $\begin{aligned}
             &[c_{00}, c_{01}, \dots, c_{0n}] \\
             &[c_{10}, c_{11}, \dots, c_{1n}] \\
             &[\dots ]
        \end{aligned}$
    };

    \node[node_style, fill=\colorA, above right = 1cm and 9.5cm of A1] (A2) {};
    \node[node_style, fill=\colorB, above right = 1cm and 9.5cm of B1] (B2) {};
    \node[node_style, fill=\colorC, above right = 1cm and 9.5cm of C1] (C2) {};
    \node[node_style, fill=\colorD, above right = 1cm and 9.5cm of D1] (D2) {};
    \node[node_style, fill=\colorE, minimum size=6mm, above right = 1cm and 9.5cm of E1] (E2) {};

    \path[-] (A2) edge node[] {} (B2);
    \path[-] (B2) edge node[] {} (C2);
    \path[-] (C2) edge node[] {} (D2);
    \path[-] (C2) edge node[] {} (E2);
    \path[-] (A2) edge[dashed, bend left=25] node[] {} (E2);

    \path[-latex, line width=2pt] (8.5 , 1.25) edge node[] {} (9.5 , 0.75);
    \path[-latex, line width=2pt] (8.5 , -1.5) edge node[] {} (9.5 , -1);
    \node[text centered] at (9, -0.12) {Prediction};
\end{tikzpicture}}
    \caption{\textbf{Learning Representative Logic Embeddings.} Some methods encode logical constraints as embeddings, then combine them with KGEs to make more informed predictions.}
    \label{fig:constraints_logic_embed}
\end{figure*}

Unfortunately, the last two approaches have the additional computational burden of grounding universally quantified rules before training. This requirement hinders their ability to scale to large KGs with many rules, a weakness reminiscent of the methods within \S\ref{subsection:fusion}. In contrast, Ding \etal \cite{ding2018improving} build upon the ComplEx algorithm, creating \textbf{ComplEx-NNE_AER}, to include non-negativity constraints on entity representations and approximate entailment constraints on relation representations. These constraints are encoded through a penalty term applied to the objective function. The authors argue that these constraints are sufficient to impose a compact, informative prior structure of the KG on the embedding space, without negatively affecting efficiency or scalability. Most importantly, their approach does not require rule grounding like the previously mentioned ones. Unfortunately, however, constraints are incorporated into the loss in a way that does not support general first-order logic rules like that of \textbf{LRLE} or \textbf{KALE}. We thus surmise that their approach might perform comparatively worse on more complex KG structures.

More generally, Ahmed \etal \cite{ahmed2022neuro} encode constraints as a loss function that can be applied to any NN model, including GNNs. In an approach termed \textbf{Neuro-Symbolic Entropy Regularisation}, the authors use \textit{semantic loss} \cite{xu-2018-semantic} as a measure of how much a first-order logic formula is satisfied by the output of a model. In contrast with methods which use fuzzy logic \cite{rocktaschel2015injecting}, the semantic loss is a probabilistic definition measuring the likelihood that the output will satisfy the constraint, given the induced probability distribution over the trained NN's outputs. The semantic loss is then combined with entropy regularization. This encourages NN predictions to conform to a structure which satisfies the specified constraints while also ensuring more distinct decision boundaries in link prediction. Similarly, \textbf{Generative KGE Circuits (GeKCs)} \cite{loconte2023turn} take advantage of probabilistic circuits \cite{vergari2020probabilistic} to ensure that hard constraints are met \cite{ahmed2022semantic}; in other words, unlike semantic loss, which encourages conformity to constraints, GeKCs guarantee that predictions will satisfy constraints.

Within this and the previous subsection, the pipeline in which KGEs are trained is linear (see Fig. \ref{fig:constraints_emb_space_or_loss_func}), with the logical constraints being injected into the pipeline at some point. Alternatively, however, logical constraints could be treated as an independent source of information to be encoded in parallel to the KG. Within the next subsection, we discuss studies that explore that possibility.

\subsubsection{\textbf{Learning Representative Logic Embeddings}}
\label{logcon_log_embed}

Within the last subcategory of learning with logical constraints, represented by the rightmost branch of Fig. \ref{fig:taxonomy}-B, some approaches work by learning representative embeddings for logical constraints. Then, they systematically combine them with the KGEs before generating predictions. This idea is illustrated in Fig. \ref{fig:constraints_logic_embed}. Similar to the previous subcategories, it helps to drive training toward established and possibly domain-specific knowledge, harnessing the \mbox{\refguide} characteristic. For example, the \textbf{Quantum Logic Empowered Embedding (QLogicE)} method \cite{chen2022qlogice} uses quantum logic \cite{birkhoff1936logic} to create logic embeddings while simultaneously computing KGEs. The scoring function for each possible triple is a weighted sum of the scores for the two embedding methods, so the overall loss is also computed as a weighted composition of the two respective losses. The weights of the scoring and loss functions determine what proportion of each embedding method to take into account. Such a tactic is used within the method, \textbf{Joint Embedding of Instances and Ontological Concepts (JOIE)} \cite{hao2019universal}. Here, two separate embedding spaces are generated for the instantiated triples and the underlying ontological structure, so a joint loss is calculated. In \textbf{JOIE}, however, they found that performance is improved even further by including a third set of embeddings on a combined graph structure comprising both instantiated triples and ontological relations. Despite improved performance, one might wonder whether \textbf{JOIE} moves away from achieving an interpretable model. 

In contrast, a study by Hamilton \etal \cite{hamilton2018embedding} combines logic and node embedding spaces in a way that the logic embeddings represent a query. Their method, \textbf{Graph Query Embeddings (GQE)}, uses geometric operations which represent logical operators to compute \textit{query embeddings}. Thereafter, the likelihood that a set of nodes satisfy a query is calculated as the cosine similarity between the query embeddings and the respective node embeddings. In a highly similar manner, the \textbf{Temporal Feature-Logic Embedding framework (TFLEX)} \cite{lin2024tflex} additionally accounts for temporal dependencies by embedding timestamps. These methods essentially perform complex logical queries on a graph, making them similar to Kautz's second category of \textsc{Symbolic[Neuro] AI}.

Similarly to the above approaches, \textbf{KALE} \cite{Guo2016JointlyEK}, mentioned previously, learns embeddings based on both the triples and logical rules. Essentially, they train a KGE algorithm while computing truth values for logical rules based on the presence of instantiated triples. Training is based on loss encompassing both triples and logical formulae. While \textbf{KALE} fits most closely into this category of logical constraints, one could also use the truth values of the logical rules as confidences. In \S\ref{subsection:learning_rules}, we discuss the possibility of assigning confidence values to rules in order to quantify relative importance. Understanding which rules are most important in generating predictions can be a valuable source of interpretability. However, the authors of \textbf{KALE} do not explore this possibility.

In general, since all of the approaches described within this category use the symbolic module to guide training of the neural module, they have the potential to replace pre-training in some circumstances, as suggested in the \mbox{\refguide} characteristic. Specifically, some studies use pre-training to steer a neural module's parameters toward a relevant context; this is typically due to a lack of sufficient data pertaining to the task at hand \cite{li2016sparseness, yao2021understanding}. In this case, the symbolic module accomplishes a similar function, guiding the neural module's parameters \textit{during} training, as opposed to \textit{before} training. 

However, these approaches are weakest with regard to \mbox{\refinterp} as none of them necessarily provide insight as to why certain predictions were made over others. In fact, one could argue that the approaches which learn logic embeddings \textit{remove} the inherent interpretability that is typically so fundamental to using logic. To fully leverage this property, the symbolic and neural components of an architecture should be combined in a way that they not only inform one another but also in a way that user-level explanations are generated. In the next section, we review a variety of approaches which aim to modify or learn new logical rules to do just that.

\subsection{Rule Learning for KGC}
\label{subsection:learning_rules}

\begin{figure*}[b!]
    \begin{minipage}{0.6\textwidth}
        \centering
        \input{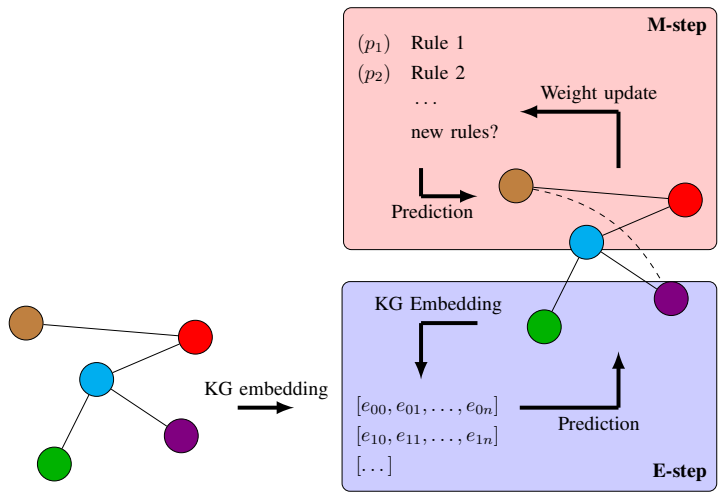}
        \scalebox{\diagramscale}{\def\colorA{brown}
\def\colorB{red}
\def\colorC{cyan}
\def\colorD{black!30!green}
\def\colorE{violet}
\def\diagramscale{0.75}
\pgfdeclarelayer{bg} 
\pgfsetlayers{bg,main} 
\begin{tikzpicture}
    \tikzset{node_style/.style={shape=circle, draw=black, minimum size=6mm}}
    
    \node[node_style, fill=\colorA] (A1) at (0, 0) {};
    \node[node_style, fill=\colorB] (B1) at (3, -0.25) {};
    \node[node_style, fill=\colorC] (C1) at (1.25, -1) {};
    \node[node_style, fill=\colorD] (D1) at (0.5, -2.5) {};
    \node[node_style, fill=\colorE] (E1) at (2.75, -2) {};

    \path[-] (A1) edge node[] {} (B1);
    \path[-] (B1) edge node[] {} (C1);
    \path[-] (C1) edge node[] {} (D1);
    \path[-] (C1) edge node[] {} (E1);

    \node[below right = 1.25cm and 5.5cm of A1, text centered, text height=0.5cm] (Emb) {
        $\begin{aligned}
             &[e_{00}, e_{01}, \dots, e_{0n}] \\
             &[e_{10}, e_{11}, \dots, e_{1n}] \\
             &[\dots ]
        \end{aligned}$
    };

    \node[above = 4.5cm of Emb, text centered, text height=1cm] (Rules) {
        $\begin{aligned}
             &(p_1) && \text{Rule 1}\\
             &(p_2) &&\text{Rule 2}\\
             & && \cdots \\ 
             & && \text{new rules?}
        \end{aligned}$
    };

    \node[node_style, fill=\colorA, above right = 2cm and 8.25cm of A1] (A2) {};
    \node[node_style, fill=\colorB, above right = 2cm and 8.25cm of B1] (B2) {};
    \node[node_style, fill=\colorC, above right = 2cm and 8.25cm of C1] (C2) {};
    \node[node_style, fill=\colorD, above right = 2cm and 8.25cm of D1] (D2) {};
    \node[node_style, fill=\colorE, above right = 2cm and 8.25cm of E1] (E2) {};

    \path[-] (A2) edge node[] {} (B2);
    \path[-] (B2) edge node[] {} (C2);
    \path[-] (C2) edge node[] {} (D2);
    \path[-] (C2) edge node[] {} (E2);
    \path[-] (A2) edge[dashed, bend left=25] node[] {} (E2);

    \path[-latex, line width=2pt] (3.75, -1.5) edge node[above] {KG embedding} (4.75, -1.5);

    \path[line width=2pt] (8.75, -1.5) edge node[below, pos=0.8] (PredictionE) {Prediction} (10.5, -1.5);
    \path[-latex, line width=2pt] (10.5, -1.5) edge node[] {} (10.5 , -0.5);

    \path[line width=2pt] (10.5, 2.75) edge node[above] {} (10.5, 3.75);
    \path[-latex, line width=2pt] (10.5, 3.75) edge node[above, pos=0.2, align=center] (Update) {Weight update} (8.75, 3.75);

    \path[line width=2pt] (7, 2.75) edge node[above] {} (7, 2.25);
    \path[-latex, line width=2pt] (7, 2.25) edge node[below, pos=0.2, align=center] (PredictionM) {Prediction} (8, 2.25);

    \path[line width=2pt] (8, 0) edge node[above, pos=0.7, align=center] (EmbeddingE) {KG Embedding} (7, 0);
    \path[-latex, line width=2pt] (7, 0) edge node[]  {} (7, -1);

    \begin{pgfonlayer}{bg}
    
    \node[] (ghostE) at (12, -0.5) {};
    \node[draw=black, fill=blue!20, opacity=0.2, rounded corners, fit=(Emb) (PredictionE) (EmbeddingE) (ghostE)] (ESTEP) {};
    \node[anchor=south east,inner sep=5pt] at (ESTEP.south east) {\textbf{E-step}};

    \node[] (ghostM) at (12, 2.25) {};
    \node[draw=black, fill=red!20, opacity=0.2, rounded corners, fit=(Rules)(Update) (PredictionM) (ghostM), inner ysep=0.4cm] (MSTEP) {};
    \node[anchor=north east,inner sep=5pt] at (MSTEP.north east) {\textbf{M-step}};
    
    \end{pgfonlayer}
\end{tikzpicture}}
    \end{minipage}%
    \begin{minipage}{0.4\textwidth}
        \caption{\textbf{EM Algorithm Methods}. These approaches adopt the idea of iterative optimization from the EM algorithm \cite{dempster1977maximum} to describe systems which iterate between symbolic and neural modules. Typically, the E-step involves KGC through a KGE method (the neural module), and the M-step involves updating the parameters of the symbolic module. To make the symbolic module (typically in the form of a MLN or rule mining approach) dynamic, some methods update rule confidences (\S\ref{learning_rule_weights}). Other methods update and alter a pool of candidate rules ({\S\ref{iterative_rule_mining}}). Note that approaches in this category may deviate from this portrayal.}
        \label{fig:em_methods}
    \end{minipage}
\end{figure*}

\noindent In the previous sections, all of the described methodologies used predefined rules for KGC. However, quite like the MLN, approaches in this final category attempt to \textit{learn} such rules. Such approaches are represented in Fig. \ref{fig:taxonomy}-C, and, collectively, they possess \textit{all} of the five characteristics listed in \S\ref{section:nesy_graphs}. Overall, most of these methods train some neural module to learn confidences for rules or systematically adjust a rule mining scheme which constitutes the symbolic module. The presence of dynamic symbolic modules makes these approaches unique from all previously described categories. In particular, it makes this category of approaches strong with regard to the \mbox{\refinterp} characteristic. As we will see next, the majority of these approaches fall into one or more subcategories based on similarities in their general architectures as well as the ways in which rules are learned.

\subsubsection{\textbf{EM Algorithm Methods}}
\label{subsection:iterative}

Many rule-learning methods claim to take an Expectation-Maximization algorithm (EM) based approach, alternating between two modules whose outcomes inform each other. These approaches are listed in the leftmost branch of Fig. \ref{fig:taxonomy}-C and distinguished by the \mbox{\refinterp}, \mbox{\refguide}, and \mbox{\refunderep} characteristics. Put succinctly, the EM algorithm is used to estimate the maximum likelihood of model parameters for situations in which there is incomplete data \cite{dempster1977maximum}. KGC, then, is an ideal and obvious setting for such methods. With varying degrees of faithfulness to the original EM algorithm proposed by Dempster \etal \cite{dempster1977maximum}, the approaches in this section all, generally, claim to adopt the iterative nature of the EM algorithm. More specifically, each approach alternates between the prediction of missing components (the ``E-step''), typically in the form of KGC, and the optimization of model parameters to account for such predictions (the ``M-step''). Normally, the parameters to be optimized belong to the symbolic module, which can be understood in this section as a dynamic framework involving logical rules (\textit{e.g.,} a MLN or rule mining system). These methods vary, however, in their interpretations of how an EM-based algorithm might be executed for KGC. 

\paragraph{\textbf{EM Methods -- Learning Rule Weights}}
\label{learning_rule_weights}

The most straightforward approach for creating a dynamic symbolic module involves assigning \textit{weights}, \textit{scores}, or \textit{confidences} to logical rules, which denote relative importance, then imposing incremental updates. Similar to approaches described in \S\ref{subsection:fusion2}, several EM methods use two complementary modules to inform one another, encapsulating the \mbox{\refguide} characteristic, but the symbolic module is now dynamic, adjusting the rule base at each iteration (Fig. \ref{fig:em_methods}). For instance, \textbf{ExpressGNN} \cite{zhang2020efficient}, \textbf{pLogicNet} \cite{qu2019probabilistic}, \textbf{pGAT} \cite{harsha2020probabilistic}, and \textbf{DiffLogic} \cite{shengyuan2023differentiable} all incorporate a MLN into the symbolic module of their algorithms to learn corresponding weights, or confidences, for logical rules. During the E-step, a KG augmented by logical inference is utilized to train a KGE method. Recall that the primary goal of the KGE method is to distill patterns from the input graph into representative embeddings, from which the KG can be reconstructed along with its missing, or latent, components. Thereafter, during the M-step, the newly predicted triples from the KGE method are used to inform and update the weights of the MLN. 

The above methods have only minor differences: for example, while \textbf{ExpressGNN} uses a vanilla GNN as the embedding method \cite{zhang2020efficient}, \textbf{pGAT} works with a variant of the GAT \cite{velivckovic2017gat} (see \S\ref{subsection:embedding_kg_completion}), utilizing the attention coefficients as an additional source of semantic information \cite{harsha2020probabilistic}. Operating similarly to \textbf{pLogicNet} and \textbf{pGAT}, Zhao \etal\!'s method, \textbf{Biomedical KG refinement with Embedding and Rules (BioGRER)} \cite{zhao2020biomedical}, is additionally specialized to operate on a biological KGs. To do so, it incorporates domain-specific knowledge into logical rules. However, unlike the aforementioned approaches, while the rule-learning module is highly similar to a MLN, it is not explicitly based on a MLN.

In contrast, Zhang \textit{et al.}'s \textbf{IterE} \cite{zhang2019iteratively} does not use a MLN. \textbf{IterE} initializes its symbolic module as a pool of randomly selected OWL-based axioms, followed by grounding. Axioms with more than one instance are selected for the final rule set. As in the previous methods, new triples inferred by the symbolic module are used as ground truth for training the neural module. Likewise, scores in the symbolic module are updated based on each iteration's KGEs. Unlike previous approaches, however, the KGEs are used directly for score updates, skipping the prediction step in between.

\paragraph{\textbf{EM Methods -- Iterative Rule Mining}}
\label{iterative_rule_mining}

While \textbf{pLogicNet}, \textbf{pGAT}, \textbf{IterE}, and \textbf{BioGRER} learn and update rule \textit{weights}, the rule set itself never changes. In contrast, Suresh and Neville's \textbf{Hybrid Method (SN-Hybrid)} \cite{suresh2020hybrid}, Lin \etal\!'s \textbf{Rule-enhanced Iterative Complementation (Rule-IC)} \cite{lin2021rule}, and Qu \etal\!'s \textbf{RNNLogic} \cite{qu2020rnnlogic} mine fresh rules over iterations, rather than using the same rules and adjusting confidences. In this sense, these methods are better aligned with the goals of ILP (\S\ref{subsection:introduction_to_logic}). In other words, KGE-derived predictions from the neural modules are used to reduce the rule mining search space and alter the pool of candidate rules, which are subsequently used for logical inference in the symbolic modules. To illustrate this, we refer the reader once again to Fig. \ref{fig:em_methods} with the notion that the rule set is iteratively altered (hence the \textit{``new rules?''} option within the rule set). Notably, \textbf{Rule-IC's} rule set is ontology-based, leading to richer KG semantics, while \textbf{SN-Hybrid} and \textbf{RNNLogic} mine general Horn rules.

Another key difference between the rule mining processes used within these approaches is the ways in which they determine the pool of candidate rules or axioms. \textbf{Rule-IC}, for example, constructs its rule set similarly to the previous section's \textbf{IterE}, but it uses a computed confidence threshold rather than a frequency threshold \cite{lin2021rule}. In \textbf{SN-Hybrid}, however, a pool of candidate rules is generated using an algorithm akin to AMIE \cite{galarraga2013amie}, in which partial rules are queued and then extended until they become proper rules. The candidate rules are later pruned via a combination of standard confidence measures, a measure of the ability to explain the existing triples, and the confidences determined through the neural module \cite{suresh2020hybrid}.

This category's essential quality of back-and-forth guidance between symbolic and neural modules makes it most similar to Kautz's third category of \textsc{Neuro;Symbolic AI}. Consequently, one major perk of EM-based algorithms is the way they account for \textit{underrepresented} relation types. Often, if a relation type occurs less frequently, KGE methods neglect to account for it, as there are fewer examples upon which to train, and the dataset imbalance creates bias toward more frequent relation types. Various KGE methods attempt to account for this via adjusted sampling or regularization methods \cite{schlichtkrull2018modeling}. Iterative methods could be used to increase the occurrence of underrepresented relation types during training. Specifically, one might consider the inference step in the symbolic modules as a way to increase the number of positive edges being fed into the embedding module. For example, if the method utilizes rule confidences, such as in \S\ref{learning_rule_weights}, one could initialize or fix rules regarding rare relation types with high confidences so that the symbolic module predicts more high-probability instances of that type. Therefore, this subcategory of approaches handles small or imbalanced datasets well, capturing the \mbox{\refunderep} characteristic. We explore this prospective direction further within \S\ref{section:prospec_direc}.

\subsubsection{\textbf{New Rules: Bridging the Discrete and Continuous Spaces}}
\label{subsection:rulebased}

In the previous subcategory, we introduced approaches with dynamic symbolic modules. Specifically, these approaches either made iterative confidence updates to predefined or mined rules (\S\ref{learning_rule_weights}) or mined a fresh set of rules based on the output of the neural module (\S\ref{iterative_rule_mining}). However, previous approaches mined and filtered rules based on predefined heuristics, including the number or proportion of groundings \cite{suresh2020hybrid} or alignment with certain OWL axioms \cite{lin2021rule}. Such heuristics may not be ideal for various applications, thereby limiting the generalizability of such methods. Alternatively, the neural module can be used directly for the task of generating and selecting descriptive rules. This leads to approaches, depicted in the centre branch of the taxonomy (Fig. \ref{fig:taxonomy}-C), that aim to learn both structural information in a \textit{discrete} space and respective weight parameters in a \textit{continuous} space \cite{yang2017differentiable}. The creation of a method that can do \textit{both} of these and train in an end-to-end fashion is inherently difficult \cite{yang2017differentiable, sadeghian2019drum}. However, the payoff is an enhanced \mbox{\refinterp} characteristic.

Methods have been developed in response to suggest new ways to perform differentiable logic-based reasoning. TensorLog \cite{cohen2016tensorlog}, for example, encodes graph entities into vectors and unique relations into respective adjacency matrices. Logical inference can then be imitated as the product of the adjacency matrices for the relations in the body of a rule with the vector of a given entity. This produces a vector in which nonzero entries represent entities for which the rule holds true. \textbf{Neural LP} \cite{yang2017differentiable} integrates TensorLog into an approach which additionally learns rule confidences in an end-to-end fashion. Specifically, it learns confidence values not just for every rule but for every relation involved in every rule, even accounting for differences in the varying lengths of rule bodies. \textbf{DRUM} \cite{sadeghian2019drum} is a refined version of \textbf{Neural LP} in which the authors point out that \textbf{Neural LP} learns high confidences for incorrect rules. To overcome this problem and reduce parameters, \textbf{DRUM} incorporates a bidirectional RNN to capture forward and backward information about possible pairings of relations, as not all relations can actually coexist in the same body of a rule. As a consequence, \textbf{DRUM} outperforms \textbf{Neural LP} and even some black-box embedding-based methods at link prediction within various datasets \cite{sadeghian2019drum}.

In contrast, the authors of \textbf{LPRules} \cite{dash2021lprules} take a slightly different approach to learning rules and confidences simultaneously. Their approach generates a weighted combination of logical rules for link prediction by iteratively augmenting a small starter pool of candidate rules. It differs from the aforementioned methods which learn confidences for the entire set of possible rules because it only works with a subset of rules at any given time, thereby reducing the search space and optimizing the time it takes to find a set of rules which are highly predictive \cite{dash2021lprules}. Another method, \textbf{RuLES} \cite{ho2018rule}, operates very similarly, augmenting its rule set iterativey by constructing and extending rules with additional atoms and logical refinement operators. \textbf{RuLES} is arguably more sophisticated, though, because the quality of candidate rules is checked against pre-computed KGEs and, optionally, text embeddings. Notably, \textbf{LPRules} and \textbf{RuLES} are comparable to the EM-based algorithms because the output of previously generated rules influences the generation of rules in the next iteration, much like that of \textbf{RNNLogic} \cite{qu2020rnnlogic}. In fact, similarly to \textbf{RNNLogic}, \textbf{LPRules} initializes its small pool of starter candidate rules via path-based heuristics. Accordingly, \textbf{RNNLogic} could also be classified within this category as it uses a NN variant for rule generation.

 Like the EM-based methods, this category is also most similar to Kautz's third classification of \textsc{Neuro;Symbolic AI}. However, the major benefit of \textbf{Neural LP}, \textbf{DRUM}, and \textbf{LPRules} is that they train end-to-end. Unlike \textbf{IterE} \cite{zhang2019iteratively} and \textbf{SN-Hybrid} \cite{suresh2020hybrid}, which use the neural module to guide rule mining, these approaches are not limited to predefined heuristics. As a major limitation, however, \textbf{Neural LP} and \textbf{DRUM} only train on positive examples and have yet to be tried on a dataset with negative examples. In many applications, positive predictions are more interesting than negative predictions. However, a lack of negative examples in the training data increases the risk of false positive predictions, which, in some contexts, such as drug-target prediction, could lead to a waste of time, money, and resources or potentially even health hazards.

\subsubsection{\textbf{Path-Based Rule Learning}}
\label{subsubsection:path_based}

\begin{figure*}[b!]
    \centering
    \input{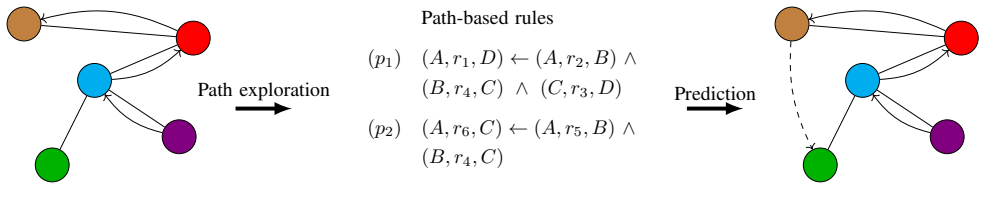}
    \scalebox{\diagramscale}{\begin{tikzpicture}
    \tikzset{node_style/.style={shape=circle, draw=black, minimum size=6mm}}
    
    \node[node_style, fill=\colorA] (A1) at (0, 0) {};
    \node[node_style, fill=\colorB] (B1) at (3, -0.25) {};
    \node[node_style, fill=\colorC] (C1) at (1.25, -1) {};
    \node[node_style, fill=\colorD] (D1) at (0.5, -2.5) {};
    \node[node_style, fill=\colorE] (E1) at (2.75, -2) {};

    \path[-] (A1) edge node[] {} (B1);
    \path[-] (B1) edge node[] {} (C1);
    \path[-] (C1) edge node[] {} (D1);
    \path[-] (C1) edge node[] {} (E1);

    \path[->] (E1) edge[bend left=20] node {} (C1);
    \path[->] (C1) edge[bend right=20] node {} (B1);
    \path[->] (B1) edge[bend right=20] node {} (A1);

    \node[below right = 1.25cm and 5.75cm of A1, text centered, text height=0cm] (O) {
        $\begin{aligned}
            &&&\text{Path-based rules}\\[0.5em]
            &(p_1) && (A, r_1, D) \leftarrow (A, r_2, B) \ \land \\
            &      && (B, r_4, C) \ \land \ (C, r_3, D) \\[0.5em]
            &(p_2) && (A, r_6, C) \leftarrow (A, r_5, B) \ \land \\
            &      && (B, r_4, C) \\[0.5em]
            & && \cdots
        \end{aligned}$
    };

    \node[node_style, fill=\colorA, right = 13cm of A1] (A2) {};
    \node[node_style, fill=\colorB, right = 13cm of B1] (B2) {};
    \node[node_style, fill=\colorC, right = 13cm of C1] (C2) {};
    \node[node_style, fill=\colorD, right = 13cm of D1] (D2) {};
    \node[node_style, fill=\colorE, minimum size=6mm, right = 13cm of E1] (E2) {};

    \path[-] (A2) edge node[] {} (B2);
    \path[-] (B2) edge node[] {} (C2);
    \path[-] (C2) edge node[] {} (D2);
    \path[-] (C2) edge node[] {} (E2);

    \path[->] (E2) edge[bend left=20] node {} (C2);
    \path[->] (C2) edge[bend right=20] node {} (B2);
    \path[->] (B2) edge[bend right=20] node {} (A2);
    \path[->] (A2) edge[dashed, bend right=15] node[sloped, anchor=center] {} (D2);

    \path[-latex, line width=2pt] (3.75, -1.5) edge node[above] {Path exploration} (4.75, -1.5);
    \path[-latex, line width=2pt] (11.75, -1.5) edge node[above] (Pred) {Prediction} (12.75, -1.5);
\end{tikzpicture}}
    \caption{\textbf{Path-Based Rule Learning.} These methods generate rules based on paths in the KG to encode long-range dependencies. Path exploration is sometimes guided by a NN or KGE method, and confidence scores are often computed for rules.}
    \label{fig:path-based}
\end{figure*}

Most of the previous approaches which mine or generate rules do so in a way that treats the latter as statements involving individual relation types. However, there exists another branch of work, the rightmost part of Fig. \ref{fig:taxonomy}-C, which accomplishes rule-learning through path-based approaches. Path-based methods can make inferences from chains of edges or relations, as shown in Fig. \ref{fig:path-based}. Consequently, the generated rule sets are not only interpretable but often more expressive, as they facilitate understanding of relationships between nodes that are several hops away from each other, leveraging long-range dependencies. Therefore, in addition to leveraging all the other listed characteristics in \S\ref{section:nesy_graphs}, this subcategory of approaches is \textit{uniquely} distinguished by the \mbox{\reflongr} characteristic.

Long-range dependency is particularly important in application domains such as biology, in which a chain of interactions between macromolecules connects entities several steps apart \cite{chang2001mammalian, crino2016mtor}. This is why Liu \etal \cite{liu2021neural} design a path-based neurosymbolic method for predicting novel drug indications (\textit{i.e.}, treatment possibilities for pharmaceutical drugs \cite{domingo2021covid}). Specifically, they investigate paths which start and end with \textit{drug} and \textit{disease} nodes, respectively. By exploring indirect drug indications which might operate through other entities, such as genes, they expand the set of novel discoveries possible. However, using path-based exploration expands the search space. Therefore, to search the KG meaningfully, they employ \textbf{Policy-guided Walks with Logical Rules (PoLo)}. Essentially, they start by encoding general path patterns, or \textit{metapaths}, as logical rules with pre-computed confidence scores. While metapaths can be manually selected from expert knowledge, they can also be mined, such as in the \textbf{PoLo} follow-up study  by Dranc\'{e} \etal \cite{drance2021neuro}. In both studies, they train an agent to walk through a biological KG, storing history using a LSTM. To use the rules as guidance, the agent is not only rewarded when it finds a positive drug indication but also when the path it follows corresponds to one of the logically encoded metapaths. Ultimately, the incorporation of path-based rules in the symbolic module guides the neural module toward exploration of long-range dependencies so that indirect relationships between distant nodes can be predicted. Therefore, this can be seen as another example of the \mbox{\refguide} characteristic. Furthermore, expert-curated metapaths tend to consider the most probable combinations of relationships between various node types, so these approaches have the potential to handle heterogeneous graphs well, fitting the \mbox{\refhetagg} characteristic. Alternatively, if one uses rule mining to generate the metapaths, as \mbox{Dranc\'{e}} \mbox{\etal}did, the generated rules and their corresponding confidences facilitate the \mbox{\refinterp} characteristic of the model.

\textbf{LPRules}, previously mentioned in \S\ref{subsection:rulebased}, operates in a similar style to \textbf{PoLo}. Specifically, one of the heuristics which \textbf{LPRules} proposes to generate and update the pool of candidate rules utilizes long-range dependencies in the KG. For rule generation regarding a specific relation, such as drug indications in the \textbf{PoLo} study, it iterates through instances of that relation and finds the shortest path between each respective pair of nodes, excluding the current direct edge. From the shortest path, a rule is generated based on its general sequence of relations (which \textbf{PoLo} called a metapath) \cite{dash2021lprules}. From the aforementioned studies, we can see that KGC can depend heavily upon information beyond a node's local neighborhood. However, one major challenge with path-based approaches is that various sequences of relations are often imbalanced, as some relations and patterns thereof are significantly rarer than others.

Dependent upon the application, less frequent path sequences might be less interesting or relevant to the end-goal. Sen \etal \cite{sen2021combining} aim to address this. In their study, they first utilize a Mixture of Paths (MP), in which the body of each generated logical rule contains one of many possible length-$k$ relation sequences in the KG. To generate such rules, they use the so-called Logical NN (LNN) \cite{riegel2020logical}, coining their neurosymbolic approach \textbf{LNN-MP}. The previously mentioned \textbf{RNNLogic} (see \S\ref{subsection:rulebased}) also generates path-based rules in a similar manner to the MP method. However, Sen \etal argue that RNN-based approaches tend to be unnecessarily complex. Because LNNs are based on real-valued boolean logic, \textbf{LNN-MP} is computationally simpler than RNN-based approaches, and the rules learned are fully interpretable. Thereafter, to represent the most common path patterns in the KG, \textbf{LNN-MP} can be used on KGEs pre-trained with bias toward more prevalent relation sequences. By doing so, \textbf{LNN-MP} is capable of learning rules most relevant to the paths present in the KG, improving KGC performance.

Similarly, an approach by Hirose \etal \cite{hirose2021transductive} weighs path-based rules on relation path frequencies in the KG, focusing KGC upon the most prevalent relation sequences. Their method, \textbf{Transductive Augmentation (TA)}, draws inspiration from \textbf{UniKER}, described in \S\ref{subsection:fusion2}. Like the latter, it follows the same general process, illustrated within Fig. \ref{fig:modular_two_step}, of iteratively augmenting a KG on which a KGE method trained. \textbf{TA} is made more sophisticated, however, by two major changes. First, relational path rules are mined from the KG using random walks. Then, based on the frequency of relation paths in the KG, confidence scores are computed to permit filtration for the top thousand rules. \textbf{TA} improves KGC performance in comparison to KGE methods alone, and it generates weighted, path-based rules which, once again, highlight the path-like patterns that are most relevant to predictions made.

While the benefits of path-based methods have already been abundantly discussed here, we note that these approaches also rely on several assumptions. First, they all operate on heterogeneous KGs containing multiple relation types; single-relation KGs may not benefit from such methods. Second, one might wonder, especially in the biological context of \textbf{PoLo}, whether these methods operate equally well on both directed and undirected graphs. Finally, approaches such as \textbf{LNN-MP} and \textbf{TA} assume that the most relevant path patterns are those which reflect the most prevalent sequences in the KG, but this may not always be the case; such approaches would lack the ability to handle underrepresented classes, therefore falling short of the \mbox{\refunderep} characteristic. On the other hand, Dranc\'{e} \etal point out that more interesting rules might be obtained by considering the goals of the application domain in which one is interested \cite{drance2021neuro}, and sometimes, underrepresented relation types are most important or interesting for KGC, a topic explored within \S\ref{section:prospec_direc}. 

Because the path-based approaches are varied in the way they operate, it is difficult to categorize them into one of Kautz's types. Since these approaches share the most similarity with those in the previous rule-learning sections as well as \S\ref{subsection:fusion2}, we believe that this section could be split between \textsc{Neuro:Symbolic $\rightarrow$ Neuro} and \textsc{Neuro;Symbolic} types. Because we focus on a specific area of neurosymbolic AI, we are untroubled by this imperfect division. As this area is still developing, we expect that future approaches will continue to refine and update both our classification and Kautz's.

\par

\newcommand{\cmark}{\ding{51}}
\newcommand{\xmark}{\ding{55}}
\newcolumntype{P}[1]{>{\raggedright\arraybackslash}p{#1}}
\renewcommand{\arraystretch}{1.1}
\begin{table*}[t!]
    \caption{Summary of Surveyed Neurosymbolic Approaches. \textbf{D} = incorporates domain knowledge, \textbf{L} = augments KG with logical inference, \textbf{C} = applies logical constraints, \textbf{W} = learns rule weights or confidences, \textbf{U} = updates candidate rule pool.}
    \label{tab:summary_table}
    \centering
    \begin{tabular}{c | l l P{7cm} l l l l l}
    & \textbf{Reference} 
    & \textbf{Year} 
    & \textbf{KGs} 
    & \textbf{D} 
    & \textbf{L} 
    & \textbf{C} 
    & \textbf{W} 
    & \textbf{U} 
    \\
    \hline 
    \multirow{7}{*}{\rotatebox[origin=c]{90}{
        \parbox[t]{2cm}{\centering 
            \textbf{Logically-Informed Embedding}}}}
    & Walking RDF and OWL \cite{alshahrani2017neuro, agibetov2018fast} 
        & 2017                    
        & {paper-specific biological KG}
        & \cmark & \cmark & \xmark & \xmark & \xmark \\
    & UniKER \cite{cheng2020uniker}         
        & 2020                    
        & {WN18RR, FB15k-237, Kinship} 
        & \xmark & \cmark & \xmark & \xmark & \xmark \\
    & RUGE \cite{guo2018knowledge}          
        & 2020                    
        & {FB15K, YAGO37} 
        & \xmark & \cmark & \xmark & \xmark & \xmark \\
    & RW-autodrive \cite{wickramarachchi2020evaluation} 
        & 2020 
        & {NuScenes, Lyft-Level5}
        & \xmark & \cmark & \xmark & \xmark & \xmark \\
    & SoLE \cite{zhang2019enhanced}
        & 2019 
        & {FB15K, DB100K}
        & \xmark & \cmark & \xmark & \xmark & \xmark \\
    & KGE* \cite{kaoudi2022towards}         
        & 2022                    
        & {DBpedia20k, LUBM}
        & \cmark & \cmark & \xmark & \xmark & \xmark \\
    & ReasonKGE \cite{jain2021improving}    
        & 2021                    
        & {DBpedia15k, LUBM3U, Yago3-10}
        & \cmark & \cmark & \xmark & \xmark & \xmark \\
    \hline 
    \multirow{14}{*}{\rotatebox[origin=c]{90}{
        \parbox[t]{3cm}{\centering
            \textbf{Logical Constraints}}}}
    & Neuro-Symbolic Entropy Reg. \cite{ahmed2022neuro} 
        & 2022 
        & {ACE05, SciERC}
        & \xmark & \xmark & \cmark & \xmark & \xmark \\
    & GeKCs \cite{loconte2023turn}
        & 2023
        & {FB15k-237, WN18RR, ogbl-biokg}
        & \cmark & \xmark & \cmark & \xmark & \xmark \\
    & ReOnto \cite{jain2023reonto}
        & 2023
        & {BioRel, ADE}
        & \cmark & \xmark & \cmark & \xmark & \xmark \\
    & R2N \cite{marra2021neural}            
        & 2021                    
        & {Countries dataset, Nations dataset, UMLS, Kinship, Cora}
        & \xmark & \xmark & \cmark & \xmark & \xmark \\
    & SLRE \cite{guo2020knowledge}          
        & 2020                    
        & {FB15K, DB100K}
        & \xmark & \xmark & \cmark & \xmark & \xmark \\
    & GCR \cite{HChenGCR}                   
        & 2022                    
        & {Amazon e-commerce, FB15k-237}
        & \xmark & \xmark & \cmark & \xmark & \xmark \\
    & QLogicE \cite{chen2022qlogice}        
        & 2022                    
        & {FB15k, FB15k-237, YAGO3-10, UMLS, Kinship, WN18RR} 
        & \xmark & \xmark & \cmark & \xmark & \xmark \\
    & ComplEx-NNE\_AER \cite{ding2018improving} 
        & 2018                
        & {FB15K, DBpedia, WN18}
        & \xmark & \xmark & \cmark & \xmark & \xmark \\
    & KALE \cite{Guo2016JointlyEK}          
        & 2016                    
        & {WN18, FB122}
        & \xmark & \xmark & \cmark & \cmark & \xmark \\
    & JOIE \cite{hao2019universal}            
        & 2019                    
        & {YAGO26K-906, DB111K-174}
        & \cmark & \xmark & \cmark & \xmark & \xmark \\
    & Graph Query Embeddings \cite{hamilton2018embedding}
        & 2018                    
        & {paper-specific biological KG, Reddit}
        & \xmark & \xmark & \cmark & \xmark & \xmark \\
    & TFLEX \cite{lin2024tflex}
        & 2024                 
        & {paper-original: ICEWS14, ICEWS05-15, GDELT-500}
        & \xmark & \xmark & \cmark & \xmark & \xmark \\
    & CQD\(\mathbf{^{\mathcal{A}}}\) \cite{cqd, cqd_a}
        & 2023                 
        & {FB15K, FB15K-237, NELL995}
        & \xmark & \xmark & \cmark & \xmark & \xmark \\
    & KeGNN \cite{werner2023knowledge}
        & 2023               
        & {Cora, Citeseer, PubMed, Flickr}
        & \xmark & \xmark & \cmark & \xmark & \xmark \\
    & Low-rank Logic Embeddings \cite{rocktaschel2015injecting}  
        & 2015 
        & {Freebase}
        & \xmark & \cmark & \cmark & \xmark & \xmark \\
    & SimplE+ \cite{fatemi2019improved}     
        & 2019                    
        & {FB15K, WN18}
        & \xmark & \xmark & \cmark & \xmark & \xmark \\
    & KGE\textsuperscript{R} \cite{minervini2017regularizing} 
        & 2017   
        & {DBpedia, YAGO, WordNet}                   
        & \xmark & \xmark & \cmark & \xmark & \xmark \\
    \hline 
    \multirow{21}{*}{\rotatebox[origin=c]{90}{
    \parbox[t]{4cm}{\centering
        \textbf{Rule Learning}}}}
    & LPRules \cite{dash2021lprules}        
        & 2021                    
        & {FB15k-237, YAGO3-10, DB111K-174, WN18RR, UMLS, Kinship}           
        & \xmark & \xmark & \xmark & \cmark & \cmark \\
    & RuLES \cite{ho2018rule}
        & 2018                    
        & {FB15K, Wiki44K, Family Dataset}           
        & \xmark & \xmark & \xmark & \cmark & \cmark \\
    & pLogicNet \cite{qu2019probabilistic}  
        & 2019                    
        & {FB15k, FB15k-237, WN18, WN18RR}                                   
        & \xmark & \cmark & \xmark & \cmark & \xmark \\
    & PoLo \cite{liu2021neural, drance2021neuro}            
        & 2021                    
        & {Hetionet, OREGANO KG}
        & \cmark & \xmark & \xmark & \cmark & \xmark \\
    & pGAT \cite{harsha2020probabilistic}   
        & 2020                    
        & {WN18RR, FB15K-237}                
        & \xmark & \cmark & \xmark & \cmark & \xmark \\
    & DiffLogic \cite{shengyuan2023differentiable}
        & 2023                
        & {CodeX, YAGO3-10, WN18, WN18RR, Kinship}              
        & \xmark & \cmark & \xmark & \cmark & \xmark \\
    & Neural LP \cite{yang2017differentiable} 
        & 2017                  
        & {WordNet18, Freebase15K, Freebase15KSelected, UMLS, Kinship, WikiMovies}
        & \xmark & \xmark & \xmark & \cmark & \cmark \\
    & RNNLogic \cite{qu2020rnnlogic}
        & 2020
        & {WN18RR, FB15k-237, UMLS, Kinship}
        & \xmark & \xmark & \xmark & \cmark & \cmark \\
    & DRUM \cite{sadeghian2019drum}
        & 2019
        & {UMLS, Kinship, Family Dataset, WN18RR, FB15K-237}
        & \xmark & \xmark & \xmark & \cmark & \cmark \\
    & ExpressGNN \cite{zhang2020efficient}  
        & 2020
        & {Cora, UW-CSE, Kinship, FB15K-237}
        & \xmark & \cmark & \xmark & \cmark & \xmark \\
    & BioGRER \cite{zhao2020biomedical}
        & 2020
        & {kg-covid-19}
        & \cmark & \xmark & \xmark & \cmark & \xmark \\
    & SN-Hybrid \cite{suresh2020hybrid} 
        & 2020
        & {YAGO3-10, FB15K-237, WN18RR}
        & \xmark & \xmark & \xmark & \cmark & \cmark \\
    & Rule-IC \cite{lin2021rule}
        & 2021
        & {FB15k, FB15k-237, WN18, WN18RR}
        & \xmark & \xmark & \xmark & \cmark & \cmark \\
    & IterE \cite{zhang2019iteratively}
        & 2019
        & {FB15k, FB15k-237, WN18, WN18RR}
        & \xmark & \xmark & \xmark & \cmark & \xmark \\
    & LNN-MP \cite{sen2021combining}
        & 2021
        & {UMLS, Kinship, WN18RR, FB15k-237}
        & \xmark & \xmark & \xmark & \cmark & \cmark \\
    & Transductive Augmentation \cite{hirose2021transductive}
        & 2021                    
        & {WN18RR, FB15k-237}                                           
        & \xmark & \cmark & \xmark & \cmark & \xmark \\
    \end{tabular}
 \end{table*}

Within the last decade and particularly within the last six years, one can see that neurosymbolic approaches for reasoning on KGs have gained significant interest. We found that the surveyed approaches fit best within three major categories: (1) logically-informed embedding approaches (\S\ref{subsection:fusion}), which are most similar to Kautz's \textsc{Neuro:Symbolic~$\rightarrow$~Neuro} category, (2) embedding approaches with logical constraints (\S\ref{subsection:logical_constraints}), most like \textsc{Neuro[Symbolic] AI}, and (3) rule-learning approaches (\S\ref{subsection:learning_rules}), which fit most closely to \textsc{Neuro;Symbolic AI}. Furthermore, we anticipate that this area of work will expand significantly within the next few years to fill out the parts of Kautz's neurosymbolic classifications which we did not frequently mention, such as \textsc{Symbolic[Neuro] AI}. Additionally, we expect that these approaches, along with their unique capabilities, will fill in several research gaps. In an attempt to facilitate and inspire future studies, we discuss prospective uses in the next section.


\section{Limitations \& Prospective Directions}

\noindent As neurosymbolic reasoning over KGs is still a young field of research, there are plenty of technical and practical areas yet to be fully explored. We next mention several common limitations of these approaches, then suggest a number of prospective directions, which we hope will cultivate a greater interest in this domain. In particular, we note that many of these directions could be useful for biomedical data and applications, so we use a number of examples in this domain to illustrate our points.

\subsection{Limitations}

While respective benefits and weaknesses were discussed for each technical category, there are also several \textit{general} limitations of these studies as well as neurosymbolic approaches as a whole.

\subsubsection{Increased Complexity} While, in theory, neurosymbolic methods could decrease computational complexity by guiding neural training, they may also \textit{increase} it, depending upon their implementation. This is because existing symbolic methods which rely upon grounding rules, like the traditional implementation of the MLN, do not tend to scale well to large datasets \cite{sun2017scalable}. Additionally, as explained in \S\ref{subsection:rulebased}, it is difficult to integrate the symbolic module with the neural module in an end-to-end fashion. Therefore, if a neurosymbolic pipeline fails to integrate the symbolic module in a way that improves scalability, then combining it with a neural module increases model complexity.
    
\subsubsection{Stacking-Induced Performance Gains} Several surveyed approaches stack two models, one each for symbolic and neural processes, into one combined model. Examples of such approaches can be found in \S\ref{subsection:fusion} and \S\ref{subsection:iterative}. Sometimes, studies using these approaches claim to see improved performance in comparison to the use of the neural module alone \cite{perozzi2014deepwalk, wickramarachchi2020evaluation}. However, work such as that by Rivas-Barragan \mbox{\etal} \cite{rivas2022ensembles} have found that \textit{ensemble methods} for KGs tend to see improved performance than the use of a single model. This calls to question whether the observed performance increase is simply due to model stacking, as opposed to the neurosymbolic aspect.
    
\subsubsection{Domain Knowledge Availability} Several neurosymbolic approaches discussed, such as those that require an ontology \cite{wickramarachchi2020evaluation, jain2023reonto}, use domain-specific knowledge. Unfortunately, expert-curated knowledge is not readily available or digitized for every field. In such cases, these approaches are not easily applicable.

\subsubsection{Ill-defined Interpretability} As discussed in \S\ref{section:nesy_graphs}, a major characteristic of the surveyed approaches is their facilitation of model interpretability. However, the concept of interpretability is generally ill-defined \cite{molnar2020interpretable}, and it is not captured consistently across the surveyed approaches. Therefore, while there is a general consensus that neurosymbolic AI can improve interpretability, there is no official metric for it, making this a subjective evaluation.

\subsection{Prospective Directions}

\subsubsection{Underexplored Application Areas}

The selection of application domains chosen in the surveyed papers was largely driven by a preference for commonly used and openly available KGs, including YAGO \cite{suchanek2007yago} and DBpedia \cite{auer2007dbpedia}, which comprise general knowledge. While this is ideal for benchmarking, few studies demonstrated their method for a specific task or within a certain domain. Now, with our proposed taxonomy at hand and a general understanding of each category's features (see Table \ref{tab:summary_table}), prospective studies could aim to do so. For instance, surveyed methods which foster increased interpretability, such as those within \S\ref{subsection:learning_rules}, are likely to be especially helpful for areas that closely affect peoples' health or livelihoods, such as biomedicine, for reasons previously established, or autonomous driving, to ensure user and bystander safety \cite{molnar2022}. Additionally, models which affect financial decisions, such as loan applications or customer turnover, might also have significant impacts on business success and benefit from interpretability \cite{molnar2022}. Alternatively, if a field is already well investigated and ample expert knowledge is available, this expertise could be used in a KG augmentation step as in \S\ref{subsection:fusion} or as training constraints as in \S\ref{subsection:logical_constraints}. Example applications of this could include tasks within the natural sciences in which there are several openly available databases of information (\textit{e.g.}, biology \cite{ashburner2000gene, schriml2022human} or chemistry \cite{gaulton2017chembl}), language-oriented tasks which rely on human-defined rules \cite{gollapalli2017incorporating, zhuang2017challenges}, and multimedia computing, which is built on human-made features \cite{zhuang2017challenges}. Because this field is still developing, there are plenty of new territories that researchers might explore.

\subsubsection{Multimodal Data Integration}

Multimodal data describes a dataset or a combination of datasets that contain various forms, such as images, videos, text or long sequences and numerical measurements. For example, a news article might contain both text and images to describe a story \cite{gao2020survey, lahat2015multimodal}. Specifically, multimodal data tends to be particularly popular within the biomedical and autonomous driving domains, among others \cite{lahat2015multimodal, yin2017automatic}. While using a single type of data at a time is much simpler, various data formats can complement and complete one another. However, fusing the various data modalities into a unified input is challenging: one must find a way to represent all modalities similarly, such as numerically, while avoiding adding bias. Existing approaches tend to either (1) concatenate modalities into one, input vector, in which modalities are no longer distinguishable, (2) project modalities into representative embdeddings via an autoencoder, or (3) use entirely separate models for each modality \cite{gao2020survey, stahlschmidt2022multimodal}. However, a neurosymbolic approach could exploit the \mbox{\refhetagg} characteristic by incorporating logic or domain-specific knowledge about the relationships between the modalities. Since multimodal data is valuable for the way its various forms complement one another, it is important to consider what kind of information these modalities contribute to one another, and how we can define these relationships through expert knowledge and logic.

\subsubsection{Conditional or Interdependent Edge Types}

In some cases, certain relations or edge types might be dependent upon one another in a way that one edge does not occur unless other edges exist. An example of this includes molecular signalling cascades, in which one protein will not interact with another unless contact is made with another protein or molecule upstream \cite{chang2001mammalian, crino2016mtor}. Such dependence could also be frequency-dependent. In another molecular signalling example, if an upstream inhibitor interacts with some protein, the less that protein may take part in other interactions \cite{gubb2010protease, rnaprotein_networks}. A similar example can be observed in traffic forecasting, in which traffic on certain roads can increase or decrease that on other roads \cite{guo2019attention, wu2022traversenet}. While such dependencies are not unique to neurosymbolic reasoning, it might be ideal for addressing them. For instance, with the incorporation of rules, neurosymbolic methods could encode dependency information between relation types. Potentially, a method such as \textbf{GCR} \cite{HChenGCR}, which uses logic to assess the probability that an edge is implied from its adjacent edges, could be ideal for interdependent edge types. Furthermore, since many neurosymbolic methods learn confidences for rules, there also exists the potential to learn confidences for which rules might coexist or influence one another. Neurosymbolic methods could, therefore, provide unique ways to interpret such dynamic KGs and the relationships between heterogeneous edge types.

\subsubsection{Spatiotemporal Reasoning}

KGs might also have spatial or temporal dependencies to consider. This is highly relevant for traffic forecasting \cite{wu2022traversenet}, biological domains \cite{kapoor2020examining}, and scene understanding \cite{cao2023discovering}. Amongst previous approaches for reasoning over spatiotemporal KGs, many of which are available through the \textit{Pytorch Geometric Temporal} package \cite{rozemberczki2021pytorch}, there is still a division between those which take rule-based approaches and those which use deep learning approaches \cite{wang2023survey}, with only one of the surveyed approaches accounting for temporal dependencies \cite{lin2024tflex}. Thus, spatiotemporal applications are a promising direction for neurosymbolic hybridization.

Similarly to the proposed approach for interdependent edge types, neurosymbolic approaches could learn rules which describe the relationships between the various edge types and time, for example. Alternatively, rules could be learned for each specific time range or spatial location. Gene expression, for instance, is the process by which our genetic code is transcribed and translated into functional products; the set of genetic codes which are expressed as well as the magnitude to which they are expressed varies completely between bodily tissue types \cite{wu2022traversenet, buccitelli2020mrnas}. One could model the interactions between the functional products as a graph, but the levels to which they affect one another depend directly upon their existence and abundance, factors controlled by the tissue type \cite{buccitelli2020mrnas}. This is an example of a network, therefore, which is spatially dependent, and learning rules specific to tissue types could unveil how processes differ between tissues. Additionally, short-term traffic forecasting, the prediction of traffic flows within a small time frame, relies heavily on spatial context, such as the layout and style of the roads being considered, as well as temporal context, such as the time of day or week \cite{lana2018road}. Such rules could add another layer of interpretability to informs researchers on how underlying processes differ across tissues, locations, time periods, and more. 

Alternatively, a neurosymbolic architecture could encode time- or space-dependent rules as constraints. For example, in a chemical or biological network, the half-lives of molecules such as messenger ribonucleic acid (mRNA) not only act as a time limit on potential interactions but also determine the levels to which various functional proteins are expressed, and therefore the extent to which those proteins can have effects on other players in the network \cite{mauger2019mrna, buccitelli2020mrnas}. Such constraints can more realistically model spatiotemporal dependencies by adding logic or domain-specific rules.

\subsubsection{Few Shot Learning}

As mentioned under the \mbox{\refunderep} characteristic, some methods increase the number of rare relation types through logical inference. Such approaches could, therefore, be used as a solution for few shot learning problems. In few shot learning, there exist a small number of instances of some class within the training data \cite{zhao2022graph}. In KGs, this could include node classes or relation types. Oftentimes, KGE methods neglect rare relation types \cite{zhang2020few}. This is often accounted for through sampling or regularization methods \cite{schlichtkrull2018modeling}, meta-learning \cite{zhang2020few, zhao2022graph}, and learned attention coefficients or relation-specific parameters, \cite{zhang2020few} but neurosymbolic methods that exploit rule-based deduction could pose an alternative solution. Methods such as \textbf{pLogicNet} \cite{qu2019probabilistic} and \textbf{pGAT} \cite{harsha2020probabilistic} augment the KG with logically inferred triples and feed the it into the neural module. The KG augmentation step may also increase the instances of rare relation types, so the neural module has a higher sample size on which to train. This poses potential to address few shot learning on KGs.

\label{section:prospec_direc}


\section{Conclusion}

\noindent Methods for reasoning on KGs are popular and widely applicable across domains \cite{zhao2020biomedical, drance2021neuro, yang2017differentiable, HChenGCR}. Therefore, it is unsurprising that there are already such a varied range of neurosymbolic methods, despite neurosymbolic AI being a young area of research. In this article, we introduce a taxonomy by which to classify these novel approaches based on the ways they contribute toward balancing interpretability, knowledge-integration, and improved predictive performance within the context of KGC. Specifically, we found that the surveyed methods fit quite well into three major categories: (1) logically-informed embedding approaches, (2) embedding approaches with logical constraints, and (3) rule-learning approaches. Throughout the article, we not only compare the various approaches but also delve into deeper subcategories of classification. For easy reference, we summarized our findings in a tabular view and compiled the available code repositories on one GitHub page\footnote{\url{https://github.com/NeSymGraphs}}. Finally, we propose prospective application-based and technical directions toward which this field might steer. Through this survey, we provide a comprehensive overview of existing methods for neurosymbolic reasoning on KGs with hopes to guide future research. 

\section*{Acknowledgments}

\noindent LND is funded by the University of Edinburgh (UoE) Informatics Graduate School through the Global Informatics Scholarship. RFM is funded by the UoE Institute for Academic Development through the Principal's Career Development PhD Scholarship. We thank these institutions for their support. We also thank Emile van Krieken, Paola Galdi, Filip Smola, Matthew Whyte, and Zonglin Ji for their thoughtful contributions.


\renewcommand{\bibfont}{\footnotesize}

\printbibliography

\vfill

\end{document}